\documentclass{article}

\usepackage[margin=1in]{geometry}

\usepackage{amsmath,amsthm,amsfonts,amssymb}
\usepackage{mathtools}
\usepackage{xcolor}
\usepackage{algorithm} 
\usepackage{algpseudocode} 
\usepackage{url}
\usepackage{multirow}
\usepackage{authblk}
\usepackage{hyperref} 
\usepackage{natbib}

\usepackage{microtype}
\usepackage{graphicx}
\usepackage{subcaption}
\usepackage{booktabs} 

\usepackage[T1]{fontenc}
\usepackage[utf8]{inputenc}
\usepackage{bbm}

\newcommand{\mybecause}[1]{&& \bigl(\because #1 \bigr)}
\newcommand{\indep}{\mathop{\perp\!\!\!\!\perp}}
\newcommand{\disteq}{\overset{\mathrm{d}}{=}}

\def\*#1{\boldsymbol{#1}}

\theoremstyle{plain}
\newtheorem*{rep@theorem}{\rep@title}
\newcommand{\newreptheorem}[2]{%
\newenvironment{rep#1}[1]{%
 \def\rep@title{#2 \ref{##1}}%
 \begin{rep@theorem}}%
 {\end{rep@theorem}}}

\newtheorem{theorem}{Theorem}[section]
\newreptheorem{theorem}{Theorem}
\newtheorem{lemma}{Lemma}[section]
\newreptheorem{lemma}{Lemma}

\newreptheorem{corollary}{Corollary}
\newtheorem{assumption}{Assumption}[section]
\newreptheorem{assumption}{Assumption}
\newtheorem{definition}{Definition}[section]
\newreptheorem{definition}{Definition}

\newcommand{\myPr}{\mathrm{Pr}}

\DeclareMathOperator*{\argmin}{arg\,min}
\DeclareMathOperator*{\argmax}{arg\,max}

\newcommand{\lw}[1]{\smash{\lower2.ex\hbox{#1}}}

\newcommand{\RR}{\mathbb{R}}
\newcommand{\NN}{\mathbb{N}}

\newcommand{\EE}{\mathbb{E}}

\newcommand{\cD}{{\cal D}}

\newcommand{\cG}{{\cal G}}

\newcommand{\cN}{{\cal N}}

\newcommand{\cP}{{\cal P}}

\newcommand{\cX}{{\cal X}}

\title{Posterior Sampling-Based Bayesian Optimization \\ with Tighter Bayesian Regret Bounds}

\author[1,2]{Shion Takeno}
\author[3]{Yu Inatsu}
\author[3]{Masayuki Karasuyama}
\author[1,2]{Ichiro Takeuchi}

\affil[1]{Nagoya University}
\affil[2]{RIKEN AIP}
\affil[3]{Nagoya Institute of Technology}

\affil[ ]{\texttt{\{takeno.shion.m6,ichiro.takeuchi.n6\}@f.mail.nagoya-u.ac.jp}}
\affil[ ]{\texttt{\{inatsu.yu,karasuyama\}@nitech.ac.jp}}
\date{}

\begin{document}
\maketitle

\begin{abstract}
    Among various acquisition functions (AFs) in Bayesian optimization (BO), Gaussian process upper confidence bound (GP-UCB) and Thompson sampling (TS) are well-known options with established theoretical properties regarding Bayesian cumulative regret (BCR).
    Recently, it has been shown that a randomized variant of GP-UCB achieves a tighter BCR bound compared with GP-UCB, which we call the tighter BCR bound for brevity.
    Inspired by this study, this paper first shows that TS achieves the tighter BCR bound.
    On the other hand, GP-UCB and TS often practically suffer from manual hyperparameter tuning and over-exploration issues, respectively.
    Therefore, we analyze yet another AF called a probability of improvement from the maximum of a sample path (PIMS).
    We show that PIMS achieves the tighter BCR bound and avoids the hyperparameter tuning, unlike GP-UCB.
    Furthermore, we demonstrate a wide range of experiments, focusing on the effectiveness of PIMS that mitigates the practical issues of GP-UCB and TS.
\end{abstract}



\section{Introduction}
\label{sec:intro}

\emph{Bayesian optimization} (BO) \citep{Mockus1978-Application} has become a popular tool for an expensive-to-evaluate black-box function optimization problem.
BO aims to optimize with fewer function evaluations by adaptively evaluating the objective function based on the Bayesian model and \emph{acquisition functions} (AF).
Due to its effectiveness, BO has been applied to a wide range of problems, including drug discovery \citep{korovina2020-chemBO}, AutoML \citep{Klein2017-Fast,Falkner2019-BOHB}, and materials informatics \citep{ueno2016combo}.

Gaussian process upper confidence bound (GP-UCB) \citep{KUSHNER1962150,Srinivas2010-Gaussian} is a seminal work of the regret analysis in BO literature.
GP-UCB achieves sub-linear Bayesian cumulative regret (BCR) bounds by sequentially evaluating the maximizer of the UCB, whose width is controlled by the confidence parameter $\beta_t$.
However, the theoretical choice of the confidence parameter can be huge.
Therefore, GP-UCB practically requires manually tuning the confidence parameter, which strongly affects the optimization performance.
Furthermore, GP-UCB with a manually tuned confidence parameter is no longer theoretically guaranteed.

Another well-known theoretically guaranteed BO method is Thompson sampling (TS) \citep{Thompson1933-likelihood}, whose same BCR bound as that of GP-UCB has been shown \citep{Russo2014-learning}.
TS is a randomized algorithm that sequentially evaluates the optimal solution of a sample path from a posterior.
Therefore, TS does not have hyperparameters like the confidence parameter for the AF, except for those for GPs.
Hence, TS does not require any manual hyperparameter tuning regarding AF.
%
%
However, it is frequently discussed that TS suffers from \emph{over-exploration}, particularly in higher-dimensional optimization problem \citep[e.g., ][]{Shahriari2016-Taking, Takeno2023-randomized}.

Recently, \citet{Takeno2023-randomized} showed improved randomized GP-UCB (IRGP-UCB) achieves tighter BCR bounds for a finite input domain compared with TS and GP-UCB.
In particular, it is shown that there is no need to increase the confidence parameter in proportion to the iterations.
However, the theoretical confidence parameter still depends on the cardinality of the finite input domain.
Since the cardinality of the finite input domain can be huge, manual tuning of the confidence parameter can be required even in IRGP-UCB.
%

\begin{table*}[t]
    \centering
    \begin{tabular}{c|c|c|c|c}
         & GP-UCB & IRGP-UCB & TS & PIMS \\ \hline
         BCR for $|\cX| < \infty$ & $O(\sqrt{T \gamma_T \log (|\cX|T)})$ & $O(\sqrt{T \gamma_T \log |\cX|})$ & * $O(\sqrt{T \gamma_T \log |\cX|})$ & * $O(\sqrt{T \gamma_T \log |\cX|})$ \\
         BCR for $\cX \subset [0, r]^d$ & $O(\sqrt{T \gamma_T \log T})$ & $O(\sqrt{T \gamma_T \log T})$ & * $O(\sqrt{T \gamma_T \log T})$ & * $O(\sqrt{T \gamma_T \log T})$
    \end{tabular}
    \caption{
        Summary of BCR bounds.
        The first and second rows show the BCR bounds for the finite and infinite input domains, respectively, where $\gamma_T$ is the maximum information gain\citep{Srinivas2010-Gaussian}, $\cX$ is the input domain, $d > 0$ is the input dimension, and $r > 0$ is a constant.
        The BCR bounds of GP-UCB and IRGP-UCB are shown in Theorem B.1 and Theorems 4.2 and 4.3 in \citet{Takeno2023-randomized}, respectively.
        %
        %
        Stars mean our results.
        }
    \label{tab:regret_summary}
\end{table*}

This study provides the same BCR bound of TS as IRGP-UCB, inspired by the analysis of IRGP-UCB, in which randomness plays a key role.
Furthermore, against the practical issues of GP-UCB-based methods and TS, we analyze a \emph{probability of improvement from a maximum of sample-path} (PIMS).
PIMS, as the name suggests, is a randomized algorithm based on the PI from the maximum of a sample path generated from the posterior.
We prove that PIMS achieves the same BCR bound as IRGP-UCB and TS.
In addition, we empirically show that PIMS alleviates the over-exploration problem of TS, although PIMS inherits the benefits of TS having no hyperparameter of AF.

One technical challenge to showing the tighter BCR bounds of TS and PIMS is handling their nature of posterior sampling-based randomization.
In the regret analyses of TS and PIMS, we consider the randomized confidence parameters, which are the same as IRGP-UCB, as a formality.
However, given the randomized confidence parameters and training dataset, although the next evaluation point $\*x_t$ in IRGP-UCB becomes constant, the next evaluation points $\*x_t$ in TS and PIMS are still random variables.
Therefore, we will show how to deal with these additional random quantities in the regret analyses.


One of the benefits of TS and PIMS is being hyperparameter-free.
%
%
Most existing analyses \citep[e.g., ][]{Srinivas2010-Gaussian,Takeno2023-randomized} have used discretization-based proofs in the BCR analysis for the continuous input domain.
Note that their discretization is only for theoretical proofs, and the algorithms can still select the next point from the original continuous space.
For TS and PIMS, to directly use a similar theoretical proof, the discretization is required not only in the proof but also in their algorithms (i.e., optimizing the sample path over finite discretized points).
However, the cardinality of the discretization must be tuned manually (e.g., through the number of grid points), which means that the AFs are no longer hyperparameter-free.
Therefore, we present an additional analysis for the BCR bound of TS and PIMS that does not require the discretization procedure in the algorithm; nevertheless, the proof is still based on the discretization.

Our contributions are summarized as follows:
\begin{enumerate}
    \item We show the sub-linear BCR bounds of TS. 
    For the finite input domain, the BCR bound is tightened compared with the known result \citep{Russo2014-learning}. 
    For the infinite input domain, we show the proof without the discretization in the algorithm.
    %
    \item We show the sub-linear BCR bounds of a BO method called PIMS, which is the PI from the maximum of the sample path from the posterior. 
    PIMS is hyperparameter-free and easy to implement, as with TS.
    For the finite input domain, PIMS achieves the tighter BCR bound as with TS and IRGP-UCB. 
    For the infinite input domain, we analyze PIMS without the discretization in the algorithm.
    %
\end{enumerate}
Theoretical results are summarized in Table~\ref{tab:regret_summary}.
Finally, we show broad experiments, particularly focusing on the practical effectiveness of PIMS compared with TS and GP-UCB-based methods.

\section{Background}
\label{sec:background}

\subsection{Problem Statement}

We consider an optimization of an unknown expensive-to-evaluate objective function $f$, which is formalized as $\*x^* = \argmax_{\*x \in \cX} f(\*x)$ using an input domain $\cX \subset \RR^d$ and an input dimension $d$.
BO aims to optimize with fewer function evaluations by adaptively observing the function value.
Thus, at each iteration $t$, we sequentially query $\*x_t$ based on the Bayesian model and AF.
We assume that the observation $y_t = f(\*x_t) + \epsilon_t$ is contaminated by a Gaussian noise $\epsilon_t \sim \cN(0, \sigma^2)$ with a positive variance $\sigma^2 > 0$.

We assume the Bayesian setting, i.e., $f$ is a sample path from a GP \citep{Rasmussen2005-Gaussian} with a zero mean and a stationary kernel function $k: \cX \times \cX \rightarrow \RR$ denoted as $f \sim \cG \cP(0, k)$.
We denote a dataset at the beginning of $t$-th iteration as $\cD_{t-1} \coloneqq \{ (\*x_i, y_i) \}_{i=1}^{n_{t-1}}$ with $n_{t-1} > 0$.
Then, the posterior distribution $p(f \mid \cD_{t-1})$ is again a GP, whose posterior mean and variance are derived as follows:
\begin{align*}
    \mu_{t-1}(\*x) &= \*k_{t-1}(\*x)^\top \bigl(\*K + \sigma^2 \*I_{n_{t-1}} \bigr)^{-1} \*y_{t-1}, \\
    \sigma_{t-1}^2 (\*x) &= k(\*x, \*x) - \*k_{t-1}(\*x) ^\top \bigl(\*K + \sigma^2 \*I_{n_{t-1}} \bigr)^{-1} \*k_{t-1}(\*x),
\end{align*}
where $\*k_{t-1}(\*x) \coloneqq \bigl( k(\*x, \*x_1), \dots, k(\*x, \*x_{n_{t-1}}) \bigr)^\top \in \RR^{n_{t-1}}$, $\*K \in \RR^{n_{t-1}\times n_{t-1}}$ is the kernel matrix whose $(i, j)$-element is $k(\*x_i, \*x_j)$, $\*I_{n_{t-1}} \in \RR^{n_{t-1}\times n_{t-1}}$ is the identity matrix, and $\*y_{n_{t-1}} \coloneqq (y_1, \dots, y_{n_{t-1}})^\top \in \RR^{n_{t-1}}$.
Hereafter, we denote that a probability density function (PDF) $p(\cdot | \cD_{t-1}) = p_t (\cdot)$, a probability $\Pr (\cdot | \cD_{t-1}) = \myPr_t (\cdot) $, and an expectation $\EE[\cdot | \cD_{t-1}] = \EE_t[\cdot]$ for brevity.

\subsection{Performance Mesuare}

This paper analyzes the \emph{Bayesian regret} \citep{Russo2014-learning}
of BO algorithms.
The BCR and Bayesian simple regret (BSR) are defined as follows:
\begin{align}
    {\rm BCR}_T &\coloneqq \EE\left[\sum_{t=1}^T f(\*x^*) - f(\*x_t) \right], \\
    {\rm BSR}_T &\coloneqq \EE \left[ f(\*x^*) - \max_{t \leq T} f(\*x_t) \right],
    \label{eq:BCR}
\end{align}
where the expectation is defined via all the randomness, i.e., $f, \{\epsilon_t\}_{t \geq 1}$, and the randomness of BO algorithms.
Since $\argmax_{t \leq T} f(\*x_t)$ is unknown in practice, the modified BSR with the recommendation, for example $\overline{\rm BSR}_T \coloneqq \EE \left[ f(\*x^*) - f(\hat{\*x}_T) \right]$, where $\hat{\*x}_T = \argmax_{\*x \in \cX} \mu_{T-1}(\*x)$, is often also analyzed \citep[Appendix~A in][]{Takeno2023-randomized}.
%

The following lemma shows the property of BSR.
\begin{lemma}
    BSR can be bounded from above as 
    $
        \overline{\rm BSR}_T 
        \leq \sum_{t=1}^T \overline{\rm BSR}_t / T
        \leq {\rm BCR}_T / T
    $
    and
    $
        {\rm BSR}_T
        \leq {\rm BCR}_T / T
    $.
    \label{lem:BSR_bound}
\end{lemma}
The proof is shown in Appendix~\ref{app:auxiliary_lemmas}.
%
%
%
%
Thus, we mainly focus on showing that BCR is sub-linear since it implies BSR converges to $0$.
In addition, for continuous input domain, we use $\overline{\rm BSR}_T \leq \sum_{t=1}^T \overline{\rm BSR}_t / T$ to show the modified BSR bound of PIMS, which is tighter than the direct consequence of BCR bound.
Furthermore, as discussed in \citet{Russo2014-learning,Takeno2023-randomized}, although the rate regarding the probability $\delta$ becomes worse, the BCR bound implies high probability bounds as a direct consequence of Markov's inequality.
Thus, although tighter regret bounds regarding $\delta$ is an important future work, we mainly discuss the BCR upper bound in this paper.


For BCR analysis, we use the following widely used assumption for the continuous $\cX$ \citep[e.g., ][]{Srinivas2010-Gaussian,Kandasamy2018-Parallelised,Takeno2023-randomized}:
\begin{assumption}
    Let $\cX \subset [0, r]^d$ be a compact and convex set, where $r > 0$.
    Assume that the kernel $k$ satisfies the following condition on the derivatives of a sample path $f$.
    There exists the constants $a \geq 1$ and $b > 0$ such that,
    \begin{align*}
        \Pr \left( \sup_{\*x \in \cX} \left| \frac{\partial f}{\partial \*x_j} \right| > L \right) \leq a \exp \left( - \left(\frac{L}{b}\right)^2 \right),\text{ for } j \in [d],
    \end{align*}
    where $[d] = \{1, \dots, d\}$.
    \label{assump:continuous_X}
\end{assumption}

Further, we define \emph{maximum information gain} (MIG) \citep{Srinivas2010-Gaussian,vakili2021-information}:
\begin{definition}[Maximum information gain]
    Let $f \sim \cG \cP (0, k)$ over $\cX \subset [0, r]^d$.
    Let $A = \{ \*a_i \}_{i=1}^T \subset \cX$.
    Let $\*f_A = \bigl(f(\*a_i) \bigr)_{i=1}^T$, $\*\epsilon_A = \bigl(\epsilon_i \bigr)_{i=1}^T$, where $\forall i, \epsilon_i \sim \cN(0, \sigma^2)$, and $\*y_A = \*f_A + \*\epsilon_A \in \RR^T$.
    Then, MIG $\gamma_T$ is defined as follows:
    \begin{align*}
        \gamma_T \coloneqq \max_{A \subset \cX; |A| = T} I(\*y_A ; \*f_A),
    \end{align*}
    where $I$ is the Shanon mutual information.
    \label{def:MIG}
\end{definition}
\noindent
It is known that MIG is sub-linear for commonly used kernel functions, e.g., $\gamma_T = O\bigl( (\log T)^{d+1} \bigr)$ for RBF kernels and $\gamma_T = O\bigl( T^{\frac{d}{2\nu + d}} (\log T)^{\frac{2\nu}{2\nu + d}} \bigr)$ for Mat\`{e}rn-$\nu$ kernels \citep{Srinivas2010-Gaussian,vakili2021-information}.

\subsection{Related Work}
\label{sec:related}

Many AFs for BO, such as expected improvement (EI) \citep{Mockus1978-Application}, entropy search (ES) \citep{Henning2012-Entropy,Villemonteix2009-aninformational}, predictive ES (PES)\citep{Hernandez2014-Predictive}, and knowledge gradients (KG)\citep{Frazier2009-knowledge}, have been proposed.
In these AFs, only the regret analysis of EI in the noiseless setting is provided  \citep{bull2011convergence}.
Therefore, particularly in the general case that noises contaminate observations, these AFs are heuristics without regret analysis.
Furthermore, although the practical effectiveness of ES, PES, and KG has been reported, they are computationally expensive and hard to implement due to cumbersome approximations.

Max-value entropy search (MES) \citep{Wang2017-Max} is a variant of ES based on the entropy of optimal value in contrast to the optimal solution used in ES and PES.
MES is easier to compute than ES and PES and empirically shows high practical performance.
\citet{Wang2017-Max} claimed that the special case of MES, in which only one Monte Carlo (MC) sample is used, achieves a high-probability simple regret bound.
PIMS is equivalent to this variant of MES with one MC sample, as shown in Lemma~3.1 in \citet{Wang2017-Max}.
%
%
However, several technical issues for the proof have been pointed out \citep{takeno2022-sequential}.
In addition, since using only one MC sample is hard to interpret in the sense of MC estimation, most experiments in \citet{Wang2017-Max} were performed with 100 MC samples.
In contrast, we provide analyses depending on the randomness of the algorithm, which implies that using only one sample is more promising.
Furthermore, MES with many MC samples requires heavy computational time.
%


Probability of improvement (PI) \citep{KUSHNER1962150,Kushner1964-new} is also a well-known BO method, which evaluates the probability that the next evaluated function value improves the best observation at iteration $t$.
To the best of our knowledge, only \citet{Wang2018-regret} have provided the high-probability bound of simple regret for PI.
However, the regret analysis in \citet{Wang2018-regret} is based on a strong assumption that the upper bound of the black-box objective function $\hat{f}^* \geq f(\*x^*)$ is known beforehand.
Although GP estimation (GP-EST) \citep{Wang2016-Optimization} can also be seen as a PI with adaptively estimated $\hat{f}^*$, its estimation guaranteeing $\hat{f}^* \geq f(\*x^*)$ is not obvious, as discussed in \citet{Takeno2023-randomized}.
%
%
If we admit Theorem~ 2 in \citet{Wang2018-regret}, it only implies that $f(\*x^*) - \max_{t \leq T} f(\*x_t) = O(\sqrt{ \gamma_T / T + \sigma^2})$, i.e., $T \bigl( f(\*x^*) - \max_{t \leq T} f(\*x_t) \bigr) = O(\sqrt{ T\gamma_T + T^2\sigma^2})$, which is linear with respect to $T$.
On the other hand, the cumulative regret $\sum_{t \leq T} f(\*x^*) - f(\*x_t) \geq T \bigl( f(\*x^*) - \max_{t \leq T} f(\*x_t) \bigr)$.
Hence, Theorem~ 2 in \citet{Wang2018-regret} does not show a sub-linear cumulative regret bound.
Thus, we believe that deriving the sub-linear cumulative regret upper bound of PI-type AFs is still an open problem.

In contrast, several kinds of research represented by GP-UCB \citep{Srinivas2010-Gaussian} and TS \citep{Russo2014-learning} achieve the sub-linear regret upper bounds.
The assumptions of the regret analysis in BO are two-fold: frequentists and Bayesian settings.
In the frequentist setting \citep[e.g., ][]{Srinivas2010-Gaussian,janz2020-bandit,Chowdhury2017-on}, $f$ is assumed to be an element of reproducing kernel Hilbert space (RKHS) specified by a kernel function for GP.
In the Bayesian setting \citep[e.g., ][]{Srinivas2010-Gaussian,Russo2014-learning,Kandasamy2018-Parallelised}, $f$ is assumed to be a sample path from the GP.
For TS, \citet{Russo2014-learning, Kandasamy2018-Parallelised} and \citet{Chowdhury2017-on} provided the regret analysis in the Bayesian and frequentists settings, respectively.
Although the regret analysis of TS and PIMS in the frequentist setting is an important research direction, this paper concentrates on the Bayesian setting.

%
Although GP-UCB achieves the sub-linear BCR bound, the theoretical $\beta_t \propto \log \bigl( |\cX| t \bigr)$ can be huge.
Thus, manually tuning confidence parameter $\beta_t$ can be a practically crucial problem.
%
%
On the other hand, \citet{Takeno2023-randomized} shows IRGP-UCB, which uses two-parameter exponential random variables $\{\zeta_t\}_{t \geq 1}$ instead of $\beta_t$, achieves the tighter BCR bound and $\EE[\zeta_t] \propto \log (|\cX|)$, in which increasing $\EE[\zeta_t]$ is not required.
However, $\log (|\cX|)$ can be large depending on the problem.
Thus, although the problem of a too-large confidence parameter is alleviated in IRGP-UCB, hyperparameter tuning is still required.
PIMS can also be interpreted as a randomized variant of GP-UCB, in which the confidence parameter is defined via the maximum of the sample path from the posterior.
See Sec.~\ref{sec:relation_PIMS_others} for details.

\section{Tighter BCR Bounds for TS}


AF of TS is the sample path from the posterior.
Algorithm \ref{alg:TS} shows an algorithm of TS.

\begin{algorithm}[t]
    \caption{TS}\label{alg:TS}
    \begin{algorithmic}[1]
        \Require Input space $\cX$, GP prior $\mu=0$ and $k$, and initial dataset  $\cD_{0}$
        \For{$t = 1, \dots$}
            \State Fit GP to $\cD_{t-1}$
            \State Generate a sample path $g_t \sim p(f | \cD_{t-1})$
            \State $\*x_t \gets \argmax_{\*x \in \cX} g_t(\*x)$
            \State Observe $y_t = f(\*x_t) + \epsilon_t$ and $\cD_{t} \gets \cD_{t-1} \cup (\*x_t, y_t)$
        \EndFor
    \end{algorithmic}
\end{algorithm}

\subsection{Regret Analysis}
\label{sec:TS_theorem}

We use the following lemma in \citet{Takeno2023-randomized}, which is a modified lemma from \citet{Srinivas2010-Gaussian}:
\begin{lemma}[Lemma~4.1 in \citep{Takeno2023-randomized}]
    Suppose that $f$ is a sample path from a GP with zero mean and a stationary kernel $k$ and $\cX$ is finite.
    Pick $\delta \in (0, 1)$ and $t \geq 1$.
    Then, for any given $\cD_{t-1}$,
    \begin{align*}
        \myPr_t \left( f(\*x) \leq \mu_{t-1}(\*x) + \beta^{1/2}_{\delta} \sigma_{t-1}(\*x), \forall \*x \in \cX \right)
        \geq 1 - \delta,
    \end{align*}
    where $\beta_{\delta} = 2 \log (|\cX| /( 2 \delta))$.
    \label{lem:bound_srinivas}
\end{lemma}

From Lemma~\ref{lem:bound_srinivas}, we show the following lemma:
\begin{lemma}
    Let $f \sim \cG \cP (0, k)$, where $k$ is a stationary kernel and $k(\*x, \*x) = 1$, and $\cX$ be finite.
    Let $\eta_t \coloneqq \frac{g_t(\*x_t) - \mu_{t-1}(\*x_t)}{\sigma_{t-1}(\*x_t)}$, where $g_t \sim p(f | \cD_{t-1})$ and $\*x_t = \argmax_{\*x \in \cX} g_t(\*x)$.
    Then, for all $t \geq 1$, the following inequality holds:
    \begin{align*}
        \EE \left[ \eta^2_t \mathbbm{1} \{ \eta_t \geq 0\} \right] 
        \leq 2 + 2 \log \bigl(|\cX|/2\bigr),
    \end{align*}
    where $\mathbbm{1}\{ \eta_t \geq 0 \} = 1$ if $\eta_t \geq 0$, and otherwise $0$.
    \label{lem:bound_TS}
\end{lemma}
See Appendix~\ref{app:TS} for the proof.
Note that $f(\*x^*)$ and $\eta_t$ do not follow Gaussian distribution since $\*x^*$ is chosen as the maximizer.


By using Lemma~\ref{lem:bound_TS}, we can obtain a tighter BCR bound, in which $\sqrt{\log T}$ factor is removed compared with  Proposition~5 in \citet{Russo2014-learning}:
\begin{theorem}
    Let $f \sim \cG \cP (0, k)$, where $k$ is a stationary kernel and $k(\*x, \*x) = 1$, and $\cX$ be finite.
    %
    %
    Then, by running TS, BCR can be bounded as follows: 
    \begin{align*}
        {\rm BCR}_T \leq \sqrt{C_1 C_2 T \gamma_T},
    \end{align*}
    where $C_1 \coloneqq 2 / \log(1 + \sigma^{-2})$ and $C_2 \coloneqq 2 + 2 \log \bigl(|\cX|/2\bigr)$. 
    \label{theo:BCR_TS_discrete}
\end{theorem}
\begin{proof}[short proof]
    For all $t \geq 1$, the following holds:
    \begin{align*}
        \EE_t \left[ f(\*x^*) - f(\*x_t) \right]
        &= \EE_t \left[ g_t(\*x_t) - f(\*x_t) \right] \\
        &= \EE_t \left[ \frac{g_t(\*x_t) - \mu_{t-1}(\*x_t)}{\sigma_{t-1}(\*x_t)} \sigma_{t-1}(\*x_t) \right] \\
        &\leq \EE_t \left[ \eta_t \mathbbm{1} \{ \eta_t \geq 0\} \sigma_{t-1}(\*x_t) \right],
    \end{align*}
    where $\eta_t = \frac{g_t(\*x_t) - \mu_{t-1}(\*x_t)}{\sigma_{t-1}(\*x_t)}$.
    Then, applying the Cauchy--Schwartz and Jensen inequalities and $\sum_{t \leq T} \sigma^2_{t-1}(\*x_t) \leq C_1 \gamma_T$ \citep{Srinivas2010-Gaussian} to the summation for all $t \leq T$, we can obtain
    \begin{align*}
        {\rm BCR}_T 
        &\leq \sqrt{ \sum_{t \leq T} \EE \left[ \eta_t^2 \mathbbm{1} \{ \eta_t \geq 0\} \right]} \sqrt{C_1 \gamma_T} \\
        &\leq \sqrt{ C_1 C_2 T \gamma_T},
    \end{align*}
    where we use Lemma~\ref{lem:bound_TS} in the second inequality.
\end{proof}
See Appendix~\ref{app:TS_discrete} for the details.
Theorem~\ref{theo:BCR_TS_discrete} shows that TS achieves the same BCR bounds as IRGP-UCB.
%
%


Next, we consider the continuous $\cX$.
In the proof for the continuous $\cX$, the discretization of $\cX$ denoted as $\cX_t$ is commonly used \citep[e.g., ][]{Srinivas2010-Gaussian}.
We describe the nearst point in $\cX_t$ of $\*x$ as $[\*x]_t$.

If we modify TS so that $[\*x_t]_t$ is evaluated instead of $\*x_t$, the same BCR upper bound of TS as that of IRGP-UCB can be obtained similarly to the existing analysis \citep{Takeno2023-randomized} (See Appendix~\ref{app:disretizedTS_continuous} for details.).
However, the theoretical discretization depends on the generally unknown parameters, such as $a$ and $b$ in Assumption~\ref{assump:continuous_X}.
Thus, discretization fineness must be set manually in the algorithm.
%
%
%
Hence, the benefit of TS having no hyperparameters is lost.
Therefore, the regret analysis without the discretization in the algorithm is also important.

\citet{Kandasamy2018-Parallelised} have shown the BCR bound of TS without the discretization for the continuous $\cX$.
However, in Eq.~(8) of \citet{Kandasamy2018-Parallelised}, non-obvious result that the $\sum_{t \leq T} \sigma^2_t([\*x_t]_t)$ is bounded from above by MIG $\gamma_T$ is used without the proof\footnote{Lemma~7 in \citep{Kandasamy2018-Parallelised} only implies that $\sum_{t \leq T} \sigma^2_t(\*x_t) \leq C_1 \gamma_T$ without the discretization.}.
Since we cannot show this non-obvious result, we show a more explicit analysis by using the following useful result \citep[Theorem~E.4 in][]{Kusakawa2022-bayesian}.
\begin{lemma}[Lipschitz constants for posterior standard deviation]
    Let $k(\*x, \*x^\prime): \RR^d \times \RR^d \to \RR$ be Linear, Gaussian, or Mat\'{e}rn kernel and $k(\*x, \*x) = 1$. 
    Moreover, assume that a noise variance $\sigma^2$ is positive.
    Then, for any $t \geq 1$ and $\cD_{t-1}$, the posterior standard deviation $\sigma_{t-1} (\*x )$ satisfies that 
    \begin{align*}
    \forall \*x,\*x^\prime \in \RR^d, \ | \sigma_{t-1} (\*x ) - \sigma_{t-1} (\*x^\prime ) | \leq L_{\sigma} \| \*x - \*x^\prime \| _1,
    \end{align*}
    where $L_{\sigma}$ is a positive constant given by 
    \begin{align*}
        L_{\sigma} = \left\{
        \begin{array}{ll}
            1 & \text{if $k(\*x,\*x^\prime)$ is the linear kernel}, \\
            \frac{\sqrt{2}}{\ell} & \text{if $k(\*x,\*x^\prime)$ is the Gaussian kernel}, \\
            \frac{\sqrt{2}}{\ell}
            \sqrt{\frac{\nu}{\nu-1} } & \text{if $k(\*x,\*x^\prime)$ is the Mat\'{e}rn kernel},
        \end{array}\right.
    \end{align*}
    where $\ell$ is length scale parameters in Gaussian and Mat\'{e}rn kernels, and $\nu$ is a degree of freedom with $\nu > 1$.
    \label{lem:Lipschitz_std_short}
\end{lemma}
For completeness, a more detailed version of this lemma is provided in Appendix~\ref{app:TS_continuous}.
Then, Lemma~\ref{lem:Lipschitz_std_short} derives the upper bound of $\sigma_t([\*x_t]_t)$:
\begin{align*}
    \sigma_t([\*x_t]_t)
    \leq \sigma_t(\*x_t) + L_{\sigma} \| \*x_t - [\*x_t]_t \|_1.
\end{align*}
Thus, appropriately refining $\cX_t$, we can obtain the following BCR bounds for continuous $\cX$ without the discretization in the algorithm:
\begin{theorem}
    Let $f \sim \cG \cP (0, k)$, where $k$ is a stationary kernel, $k(\*x, \*x) = 1$, and Assumption~\ref{assump:continuous_X} holds.
    Let $L = \max \bigl\{L_\sigma, b \bigl(\sqrt{\log (ad)} + \sqrt{\pi} / 2 \bigr) \bigr\}$, where $L_\sigma$ is defined as in Lemma~\ref{lem:Lipschitz_std_short}.
    Then, by running TS, BCR can be bounded as follows: 
    \begin{align*}
        {\rm BCR}_T \leq \frac{\pi^2}{3} + \frac{\pi^2}{6} \sqrt{s_T} + \sqrt{ C_1 \gamma_T T s_T},
    \end{align*}
    where $C_1 \coloneqq 2 / \log(1 + \sigma^{-2})$ and $s_t = 2 - 2 \log 2 + 2d \log(\lceil drL t^2 \rceil)$. 
    \label{theo:BCR_TS_continuous}
\end{theorem}
See Appendix~\ref{app:TS_continuous} for the proof.
Main differences from \citet{Kandasamy2018-Parallelised} are (i) constant factors in $s_t$ are essentially tightened as $\log(|\cX_t|)$ from $\log(|\cX_t|t^2)$ using Lemma~\ref{lem:bound_TS} (ii) the rate regarding $a$ is improved as $\log \bigl( \log a \bigr)$ from $\log a$ using Lemma~H.1 of \citet{Takeno2023-randomized} (iii) the discretization error regarding posterior standard deviation is considered as discussed earlier.

\section{Tighter Bayesian Regret Bounds for PIMS}


AF of PIMS is the PI from the maximum of the sample path from the posterior:
\begin{align*}
    \*x_t = \argmax_{\*x \in \cX} \left\{ 1 - \Phi\left( \frac{g^*_t - \mu_{t-1}(\*x)}{\sigma_{t-1}(\*x)} \right) \right\},
\end{align*}
where $g^*_t = \max_{\*x \in \cX} g_t(\*x)$, $g_t \sim p(f | \cD_{t-1})$, and $\Phi$ is the cumulative distribution function of a standard Gaussian distribution.
From the monotonicity of $\Phi$, we can rewrite as follows:
\begin{align*}
    \*x_t = \argmin_{\*x \in \cX}\frac{g^*_t - \mu_{t-1}(\*x)}{\sigma_{t-1}(\*x)}.
\end{align*}
We employ this equivalent expression due to its simplicity.
Algorithm \ref{alg:PIMS} shows an algorithm of PIMS.

Usual PI is based on the current best observation $\max_{i \leq t-1} y_i$ instead of $g^*_t$ and is known to result in over-exploitation frequently.
In the small noise variance regime, $g^*_t$ is often larger than $\max_{i \leq t-1} y_i$ and randomly generated.
Thus, PIMS explores appropriately compared with the usual PI since the AF of PI tends to explore when the base value (i.e., $\max_{i \leq t-1} y_i$ or $g^*_t$) is large.

\begin{algorithm}[!t]
    \caption{PIMS}\label{alg:PIMS}
    \begin{algorithmic}[1]
        \Require Input space $\cX$, GP prior $\mu=0$ and $k$, and initial dataset $\cD_{0}$
        \For{$t = 1, \dots$}
            \State Fit GP to $\cD_{t-1}$
            \State Generate a sample path $g_t \sim p(f | \cD_{t-1})$
            \State $g^*_t \gets \max_{\*x \in \cX} g_t$
            \State $\*x_t \gets \argmin_{\*x \in \cX} \frac{g^*_t - \mu_{t-1}(\*x)}{\sigma_{t-1}(\*x)}$
            \State Observe $y_t = f(\*x_t) + \epsilon_t$ and $\cD_{t} \gets \cD_{t-1} \cup (\*x_t, y_t)$
        \EndFor
    \end{algorithmic}
\end{algorithm}

\subsection{Regret Analysis}
\label{sec:PIMS_theorem}

First, we show a similar lemma as Lemma~\ref{lem:bound_TS} as follows:
\begin{lemma}
    Assume the same condition as in Lemma~\ref{lem:bound_TS}.
    Let $\xi_t \coloneqq \min_{\*x \in \cX} \frac{g^*_t - \mu_{t-1}(\*x)}{\sigma_{t-1}(\*x)}$.
    Then, the following inequality holds:
    \begin{align*}
        \EE \left[ \xi^2_t \mathbbm{1} \{ \xi_t \geq 0\} \right] 
        \leq 2 + 2 \log \bigl(|\cX|/2\bigr),
    \end{align*}
    for all $t \geq 1$.
    \label{lem:bound_PIMS}
\end{lemma}
See Appendix~\ref{app:PIMS_discrete} for the proof.

Using Lemma~\ref{lem:bound_PIMS}, we can obtain the following BCR bound for the finite input domain:
\begin{theorem}
    Assume the same condition as in Theorem~\ref{theo:BCR_TS_discrete}.
    %
    %
    Then, by running PIMS, BCR can be bounded as follows: 
    \begin{align*}
        {\rm BCR}_T \leq \sqrt{C_1 C_2 T \gamma_T},
    \end{align*}
    where $C_1 \coloneqq 2 / \log(1 + \sigma^{-2})$ and $C_2 \coloneqq 2 + 2 \log \bigl(|\cX|/2\bigr)$. 
    \label{theo:BCR_PIMS_discrete}
\end{theorem}
See Appendix~\ref{app:PIMS_discrete} for the proof.
Therefore, PIMS achieves the same BCR bound as TS and IRGP-UCB.


Next, we consider the continuous $\cX$.
As with TS, if we modify PIMS so that $\tilde{g}^*_t \coloneqq \max_{\*x \in \cX_t} g_t(\*x)$ is used instead of $g^*_t$, the proof of the sub-linear BCR bound of PIMS is relatively easy (See Appendix~\ref{app:disretizedPIMS_continuous} for details).
However, for the same reason as TS, the theoretical discretization in the algorithm is not preferable.
Therefore, we show the BCR bound using $g^*_t$ without the discretization in the algorithm.

On the other hand, the same proof as TS does not provide the BCR bound of PIMS.
This is because $\*x_t$ can be far away from $\argmin_{\*x \in \cX} \bigl( \tilde{g}^*_t - \mu_{t-1}(\*x) \bigr) / \sigma_{t-1}(\*x)$ in contrast to TS.
Thus, instead of Lipshitz continuity of $\sigma_{t-1}(\*x)$ (Lemma~\ref{lem:Lipschitz_std_short}), we use the following lemma, which is repeatedly used in the literature \citep[e.g., Lemma~13 in][]{Mutny2018-efficient,wang2014theoretical}:
\begin{lemma}
    Let $k$ be a kernel s.t. $k(\*x, \*x) = 1$.
    %
    Then, the posterior variance is bounded from below as,
    \begin{align}
        \sigma^2_t(\*x) \geq \frac{\sigma^2}{\sigma^2 + n_t},
    \end{align}
    for all $\*x \in \cX$ and for all $t \geq 0$, where $n_t = |\cD_t|$.
    \label{lem:LB_posterior_variance}
\end{lemma}
%
%
For completeness, the proof is shown in Appendix~\ref{app:auxiliary_lemmas}.

Then, we can obtain the following BCR bound:
\begin{theorem}
    Let $f \sim \cG \cP (0, k)$, where $k$ is a stationary kernel, $k(\*x, \*x) = 1$, and Assumption~\ref{assump:continuous_X} holds.
    Then, by running PIMS, BCR can be bounded as follows:
    \begin{align*}
        {\rm BCR}_T \leq \frac{\pi^2}{6} + \sqrt{ C_1 T \gamma_T m_T},
    \end{align*}
    where $C_1 \coloneqq 2 / \log(1 + \sigma^{-2})$ and $m_t \coloneqq 2d \log \left( \left\lceil \! t^2 b d r \bigl( \log (ad) + \sqrt{\pi} / 2 \bigr) \sqrt{(\sigma^2 + n_t) / \sigma^2} \! \right\rceil \right) - 2 \log 2 + 2$.
    \label{theo:BCR_PIMS_continuous}
\end{theorem}
See Appendix~\ref{app:PIMS_continuous} for the proof.
Since $m_t = O(\log t)$, ${\rm BCR}_T = O( \sqrt{T \gamma_T \log T})$, which is the same rate as the existing results of GP-UCB, IRGP-UCB, and TS.

As discussed in Sec.~\ref{sec:background}, our BCR bound directly implies that BSR converges to 0.
On the other hand, for the modified BSR, we can avoid the use of Lemma~\ref{lem:LB_posterior_variance} using the inequality $\overline{\rm BSR}_T \leq \sum_{t=1}^T \overline{\rm BSR}_t / T$ in Lemma~\ref{lem:BSR_bound}:
\begin{theorem}
    Assume the same condition as in Theorem~\ref{theo:BCR_PIMS_continuous}.
    Then, by running PIMS, BSR can be bounded as follows:
    \begin{align*}
        \overline{\rm BSR}_T \leq \frac{\pi^2}{6 T} + \sqrt{ \frac{C_1 \gamma_T m_T}{T}},
    \end{align*}
    where $C_1 \coloneqq 2 / \log(1 + \sigma^{-2})$ and $m_t \coloneqq 2d \log \left( \left\lceil \! t^2 b d r \bigl( \log (ad) + \sqrt{\pi} / 2 \bigr) \right\rceil \right) - 2 \log 2 + 2$.
    \label{theo:BSR_PIMS_continuous}
\end{theorem}
See Appendix~\ref{app:PIMS_continuous} for the proof.
Therefore, we can remove the term $ \log((\sigma^2 + n_t) / \sigma^2)$ in $m_t$.

\subsection{Relationship with GP-UCB}
\label{sec:relation_PIMS_others}


\citet{Wang2017-Max} have also shown the equivalence between PIMS and GP-UCB with $\beta^{1/2}_t = \xi_t$ defined in Lemma~\ref{lem:bound_PIMS}.
(We employ this definition of $\xi_t$ since $\xi_t$ can be negative.)
Details are shown in Appendix~\ref{app:equivalence}.
Therefore, PIMS can also be interpreted as a random variant of GP-UCB, whose confidence parameter is determined via $g^*_t$.
Thus, we can expect that PIMS defines the more practical confidence parameter $\xi_t$ based on the GP model in contrast to GP-UCB and IRGP-UCB, whose theoretical confidence parameters are defined by the worst-case analysis and are too conservative.
We will show this important benefit of PIMS compared with GP-UCB and IRGP-UCB in Sec.~\ref{sec:experiment}.
\section{Experiments}
\label{sec:experiment}

\begin{figure*}[t]
    \centering
    \begin{subfigure}[b]{0.3\linewidth}
        \centering
        \includegraphics[width=\linewidth]{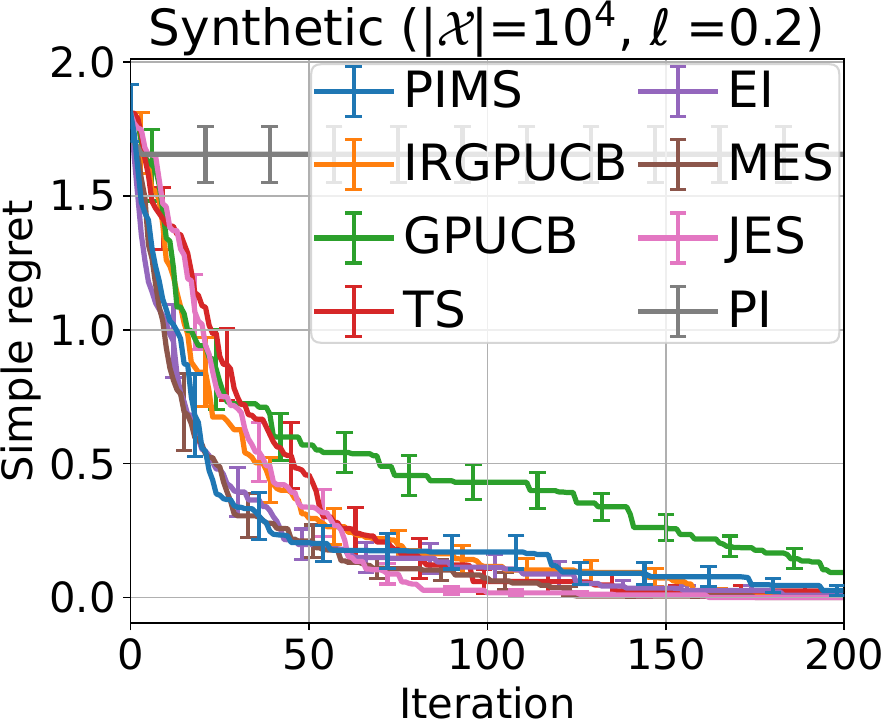}
        \includegraphics[width=\linewidth]{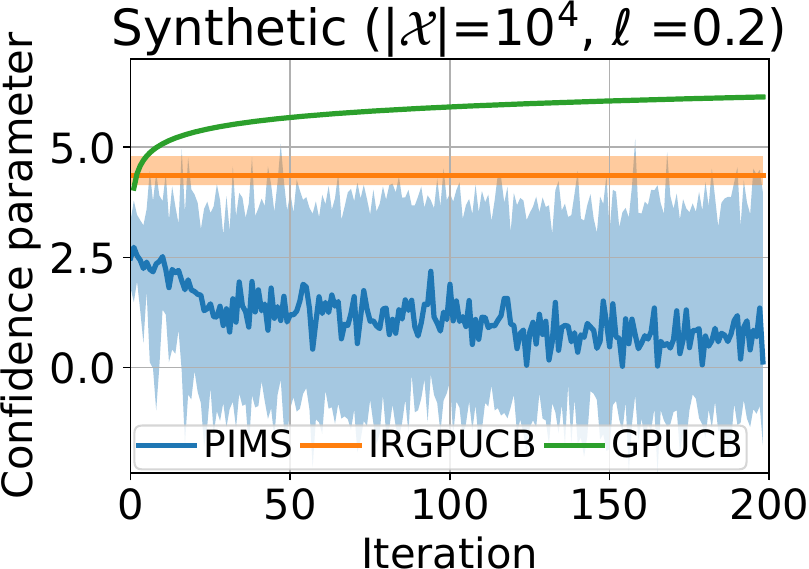}
        \caption{Default setting.}
        \label{fig:syn_reg_default}
    \end{subfigure}
    \begin{subfigure}[b]{0.3\linewidth}
        \centering
        \includegraphics[width=\linewidth]{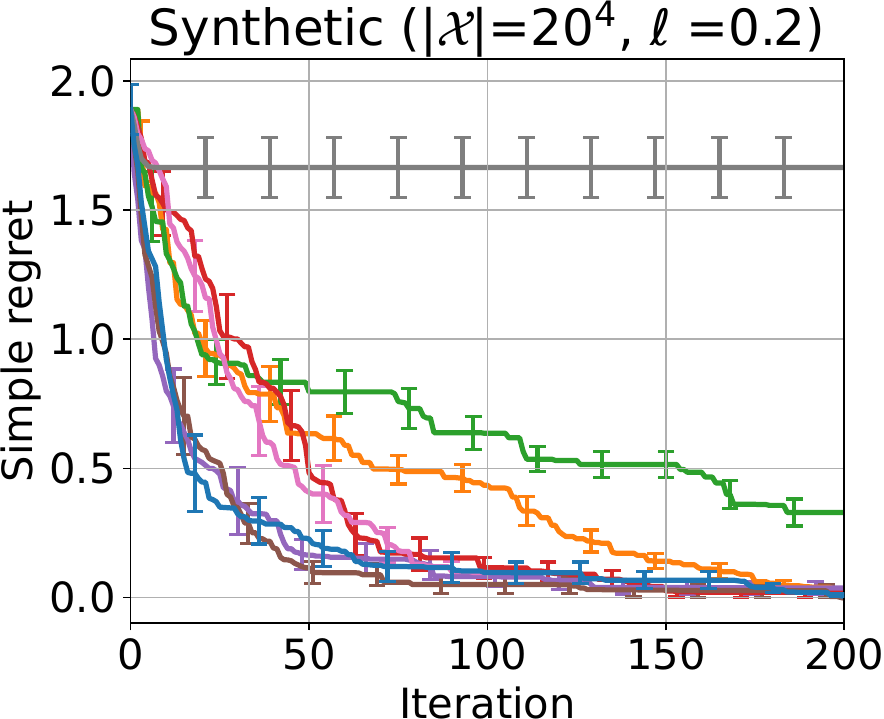}
        \includegraphics[width=\linewidth]{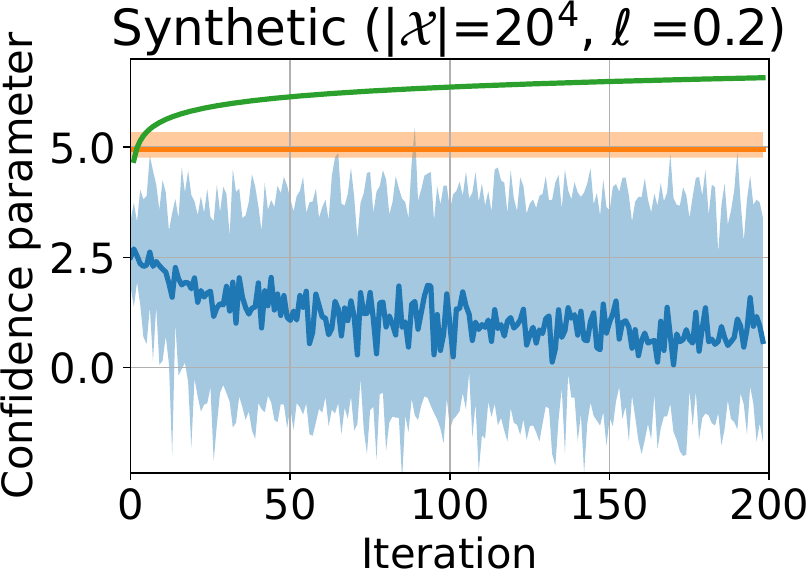}
        \caption{Large $|\cX|=20^4$.}
        \label{fig:syn_reg_largeX}
    \end{subfigure}
    \begin{subfigure}[b]{0.3\linewidth}
        \centering
        \includegraphics[width=\linewidth]{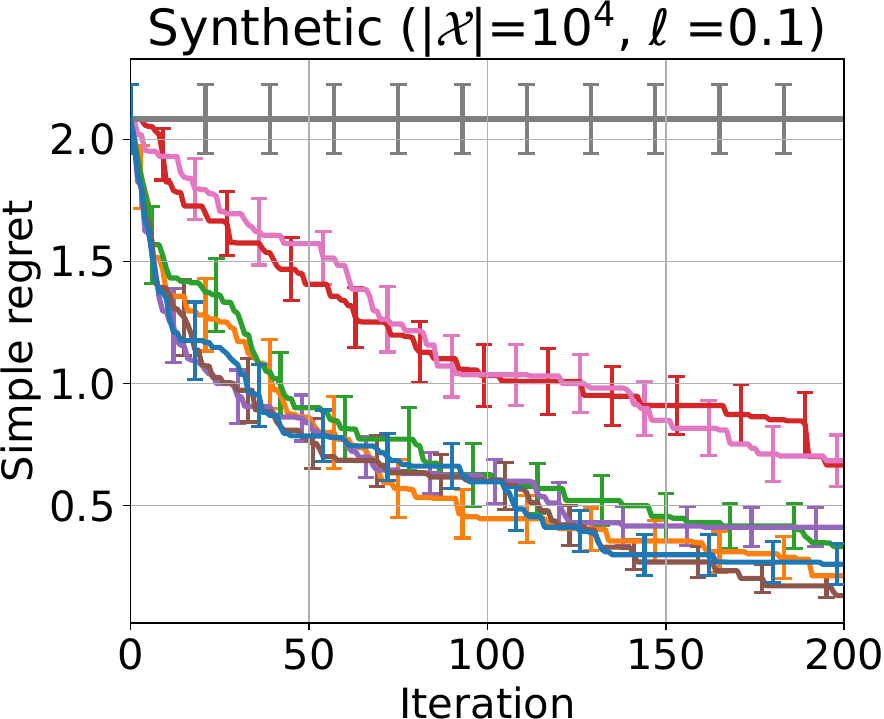}
        \includegraphics[width=\linewidth]{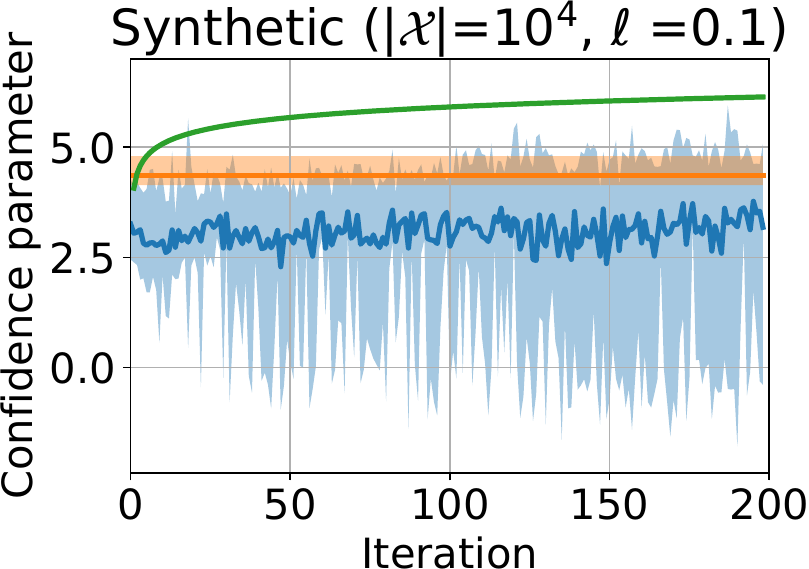}
        \caption{Small $\ell=0.1$.}
        \label{fig:syn_reg_smallell}
    \end{subfigure}
    \caption{
        The results on synthetic function experiments.
        The top row shows the average and standard error of the simple regret.
        In all the settings (a-c), we can confirm that several BO methods, including PIMS (blue), achieve the best convergence.
        In contrast, other theoretically guaranteed methods (GP-UCB, IRGP-UCB, TS) deteriorated in a certain setting.
        Therefore, we can observe that PIMS flexibly deals with various problem settings while keeping the theoretical guarantee and has superior or comparable performance compared with baselines, including heuristic methods without the theoretical guarantee, such as EI, MES, and JES.
        The bottom row represents the expectation and quantiles of $\beta_t^{1/2}$, $\zeta_t^{1/2}$, and $\xi_t$.
        }
    \label{fig:syn_regret}
    \vspace{-5pt}
\end{figure*}

We demonstrate the experiments on synthetic functions generated from GP, benchmark functions, and emulators derived from real-world datasets.
We performed the existing BO methods, EI\citep{Mockus1978-Application}, GP-UCB\citep{Srinivas2010-Gaussian}, IRGP-UCB\citep{Takeno2023-randomized}, TS\citep{Russo2014-learning}, MES\citep{Wang2017-Max}, joint entropy search (JES) \citep{hvarfner2022-joint} as baselines.
We generated $5$ inputs for the initial dataset using the Latin hypercube sampling\citep{Loh1996-Latin}.
We fixed the noise variance $\sigma^2 = 10^{-6}$ and used the RBF kernel $k(\*x, \*x^\prime) = \exp \left( - \| \*x - \*x^\prime \|_2^2 / (2 \ell^2) \right)$, where $\ell$ is hyperparameter.
%
%
%
For the posterior sampling in TS, PIMS, MES, and JES, we used the random Fourier feature (RFF) \citep{Rahimi2008-Random}.
For MC estimation in MES and JES, we generated $10$ MC samples.
Unless otherwise noted, we report the average and standard errors of the simple regret $r_t \coloneqq f(\*x^*) - \max_{t \leq T} f(\*x_t)$ over 20 random trials.

Overall, we focus on the effectiveness of theoretical BO methods with regret analysis.
In synthetic function experiments, we consider the purely theoretical setting, in which we set hyperparameter $\ell$ as one used to generate the objective function.
We aim to demonstrate that PIMS shows at least comparable performance to baselines including not theoretically guaranteed methods, in contrast to that GP-UCB-based methods and TS are degraded by over-exploration.
In benchmark function and real-world emulator experiments, we performed marginal likelihood maximization \citep{Rasmussen2005-Gaussian} every $5$ iteration.
Furthermore, in these experiments, we use the manually tuned confidence parameters for GP-UCB-based methods since theoretical confidence parameters contain unknown parameters such as $a$ and $b$.
Thus, GP-UCB-based methods are no longer theoretically guaranteed.
Hence, only TS and PIMS match the theoretical algorithm.

%

\subsection{Experiments on Synthetic Functions}

We performed the experiments for the GP-derived synthetic function and the finite input domain $\cX$.
Note that the objective function is also randomly generated for all 20 trials.
We show the results of changing the length scale parameter $\ell$ and the cardinality of $\cX$.
In particular, note that small $\ell$ implies that the objective function and the GP model are more complex.
The input domain is set as $\{ 0.1, 0.2, \dots, 1 \}^4$ when $|\cX|=10^4$ and $\{ 0.05, 0.1, \dots, 1 \}^4$ when $|\cX|=20^4$, respectively.
For this experiment, we used the theoretical confidence parameters, $\beta_t = 2 \log \left( |\cX| t^2 / \sqrt{2 \pi} \right)$ for GP-UCB and $\zeta_t \sim {\rm Exp} \left(2 \log \left( |\cX| / 2\right), 1/2 \right)$ for IRGP-UCB.

The top row of Fig.~\ref{fig:syn_regret} shows the simple regret.
In this experiment, we performed PI~\citep{Kushner1964-new}, by which we can confirm the over-exploitation tendency of the usual PI.
In contrast, we see that PIMS shows the best or comparable performance in all the plots of Fig.~\ref{fig:syn_regret} compared with the baselines, including the heuristic methods without theoretical guarantees, such as EI, MES, and JES.
Consequently, the effectiveness of PIMS under different $|\cX|$ and $\ell$ can be confirmed.

\begin{table}[t]
    \caption{Average and standard deviation of the mean of evaluated posterior standard deviation $\sum_{t \leq T} \sigma_{t-1}(\*x_t) / T$ for synthetic function experiments.}
    \centering
    \begin{tabular}{c|c|c|c}
         & Fig.~\ref{fig:syn_reg_default} & Fig.~\ref{fig:syn_reg_largeX} & Fig.~\ref{fig:syn_reg_smallell} \\ \hline
         PIMS & 0.27 $\pm$ 0.14 & 0.26 $\pm$ 0.13 & 0.71 $\pm$ 0.13 \\ 
         TS & 0.36 $\pm$ 0.14 & 0.36 $\pm$ 0.14 & 0.92 $\pm$ 0.09
    \end{tabular}
    \label{tab:posterior_var}
    \vspace{-10pt}
\end{table}

%
GP-UCB-based methods deteriorated due to the over-exploration, particularly in Fig.~\ref{fig:syn_reg_largeX} with $|\cX| = 20^4$, since the confidence parameters of GP-UCB and IRGP-UCB depend on $\cX$ as $\log |\cX|$.
This can also be confirmed by the plot of the confidence parameters shown in the bottom row of Fig.~\ref{fig:syn_regret}.
Note that PIMS can be interpreted as GP-UCB using $\xi_t$ instead of $\beta_t^{1/2}$ as discussed in Sec.~\ref{sec:relation_PIMS_others}.
Then, we can observe that $\xi_t$ is substantially smaller than $\beta_t^{1/2}$ and $\zeta_t^{1/2}$, which is conservatively set via $|\cX|$.
%
%
%
In addition, comparing Figs.~\ref{fig:syn_reg_default} and \ref{fig:syn_reg_smallell}, we can see that $\xi_t$ becomes large (exploration is emphasized) when $\ell$ is small (the objective function has many local maxima).
These results suggest that PIMS can use more practical confidence parameters based on the posterior sampling.



TS shows poor performance in Fig.~\ref{fig:syn_reg_smallell} with small $\ell$.
We conjecture that this poor performance is caused by over-exploration.
Thus, we show the average and standard deviation of the evaluated posterior standard deviation mean $\sum_{t \leq T} \sigma_{t-1}(\*x_t) / T$ for PIMS and TS in Table~\ref{tab:posterior_var}.
We can observe that TS evaluates the input whose posterior variance is larger than that of PIMS, particularly in Fig.~\ref{fig:syn_reg_smallell}.
Therefore, we can confirm that TS results in over-exploration.
It is worth noting that $\ell = 0.1$ is ideal in the sense that $\ell = 0.1$ is used to generate the objective function.
Considering the existing studies \citep[e.g., ][]{Shahriari2016-Taking}, in which TS deteriorated in the higher-dimensional problem, TS is empirically sensitive to the complexity of the GP model depending on the parameters such as $d$ and $\ell$.

\subsection{Experiments on Benchmark Functions}

\begin{figure}[t]
    \centering
    \includegraphics[height=145pt]{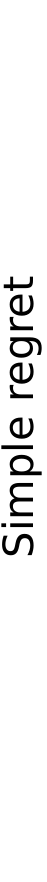}
    \includegraphics[height=145pt]{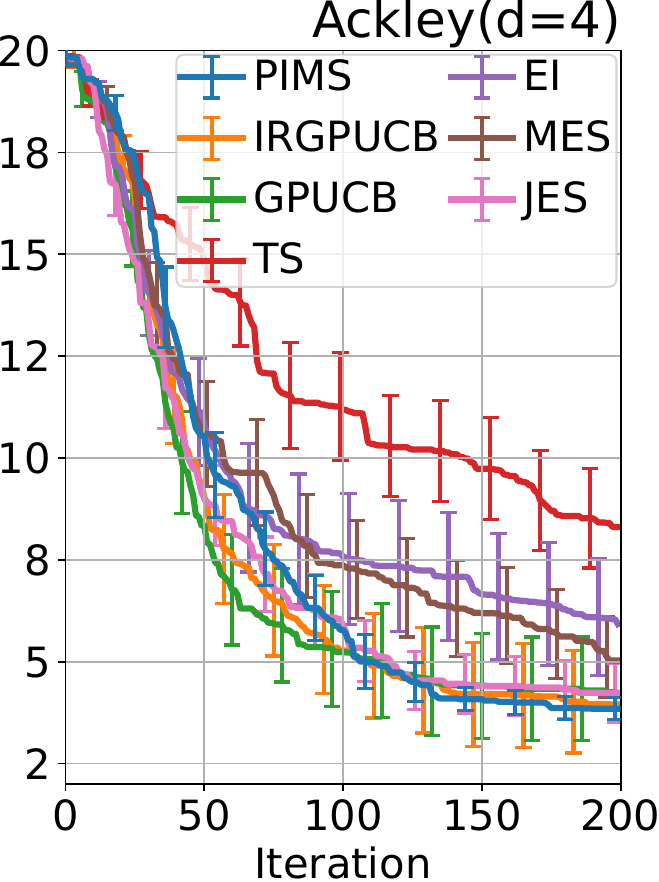}
    \includegraphics[height=145pt]{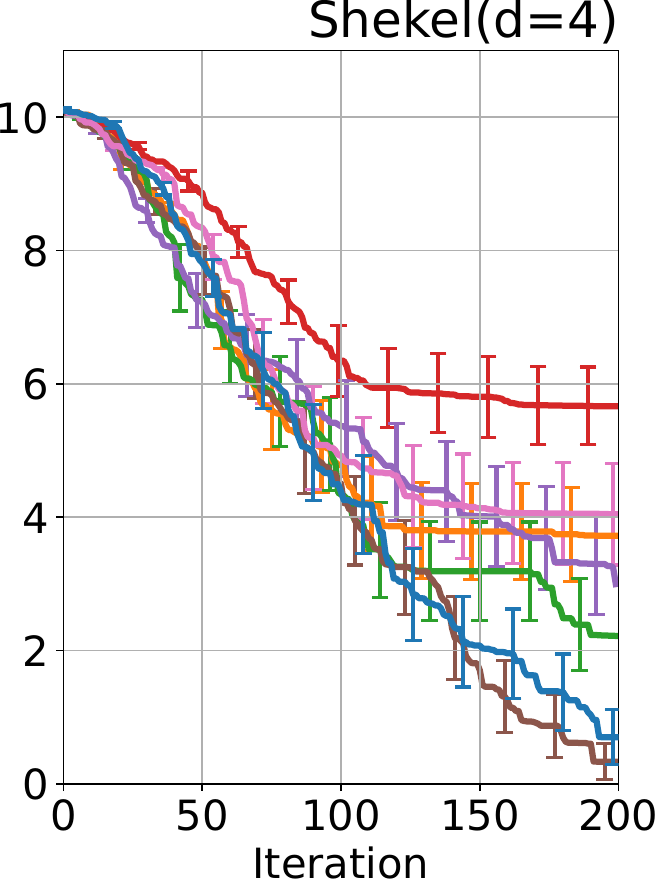}
    \caption{Average and standard error of the simple regret in benchmark function experiments.}
    \label{fig:bench_reg}
\end{figure}

Figure~\ref{fig:bench_reg} shows the results on the benchmark functions called Ackley and Shekel functions in \url{https://www.sfu.ca/~ssurjano/optimization.html}.
In this experiment, we employed the heuristic confidence parameters for GP-UCB and IRGP-UCB as  $\beta_t = 0.2 d \log \left( 2t \right)$ \citep{kandasamy2015-high} and $\zeta_t \sim {\rm Exp} \left(2 /d, 1/2 \right)$ \citep{Takeno2023-randomized}.
TS is inferior to other methods due to the over-exploration since a small $\ell$ is required to represent both functions, which have many local maxima.
On the other hand, PIMS shows comparable or superior performance in both functions compared with all the baselines, including heuristics such as EI, MES, and JES.
In particular, in the Shekel function, PIMS shows faster convergence compared with GP-UCB-based methods with the heuristic confidence parameters.
Thus, we conclude that PIMS appropriately controls the exploitation and exploration trade-off via the posterior sampling in this experiment.

\subsection{Experiments on Real-World Emulators}

We performed the experiments on the real-world data emulators \citep{hase2021olympus}.
The alkox dataset \citep{hase2021olympus} is the measurement of alkoxylation reaction with respect to 4 parameters: catalase, horseradish peroxidase, alcohol oxidase, and pH.
The Fullerenes dataset \citep{Walker2017-tuning} is the mole fraction of the desired products regarding o-xylenyl adducts of Buckminsterfullerenes with respect to 3 parameters: temperature, reaction time, and the ratio of sultine to $\rm C_{60}$.
We optimized the Bayes neural network trained by each dataset as proposed by \citet{hase2021olympus}.
Details are shown in \citet{hase2021olympus}.

\begin{figure}[t]
    \centering
    \includegraphics[height=145pt]{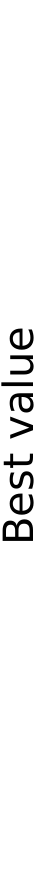}
    \includegraphics[height=145pt]{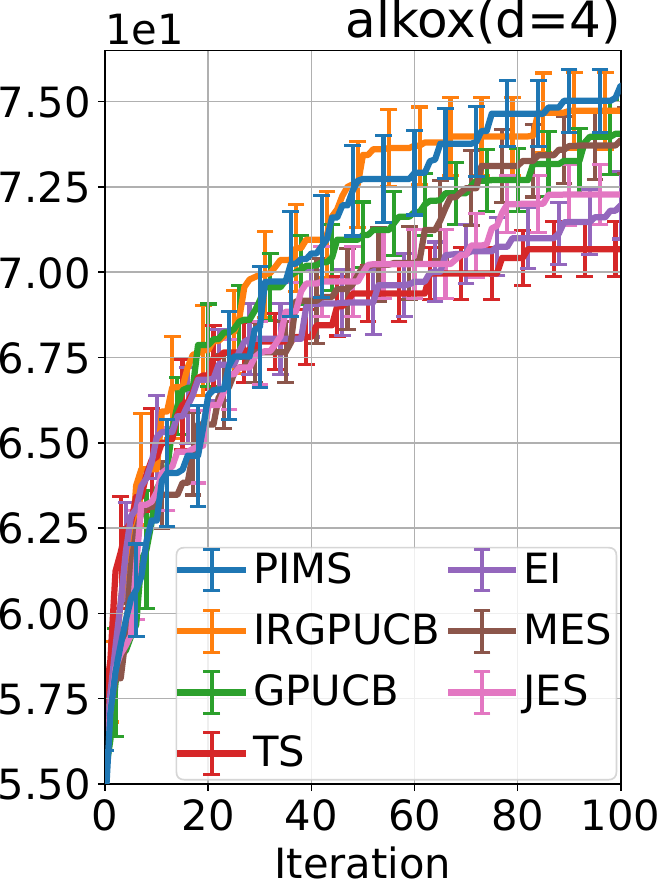}
    \includegraphics[height=145pt]{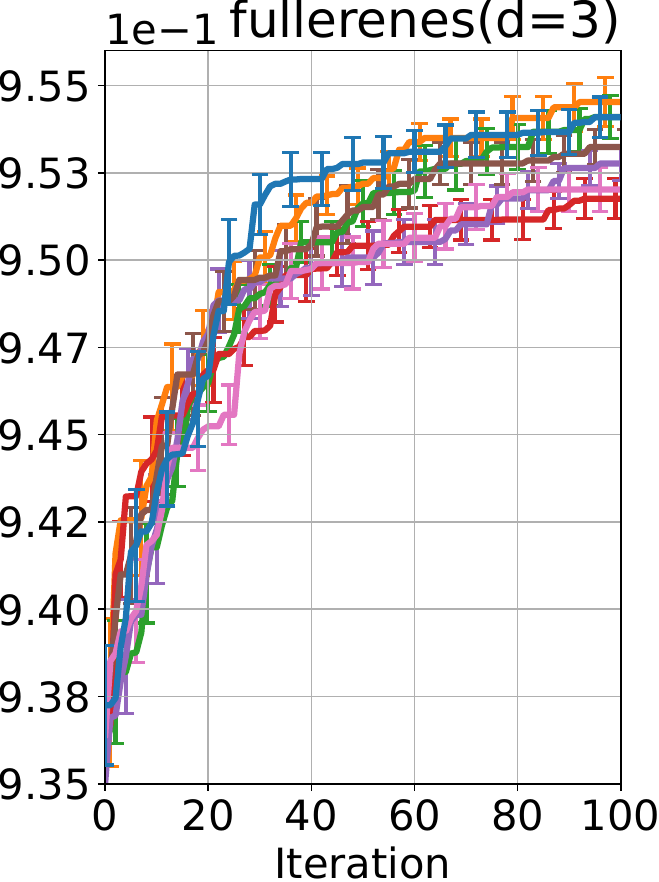}
    \caption{Average and standard error of the obtained best value in real-world emulator experiments.}
    \label{fig:real_reg}
\end{figure}

Figure~\ref{fig:real_reg} shows the best observed value since the optimal value is unknown.
Note that GP-UCB-based methods are based on the heuristic confidence parameters as with the benchmark function experiments.
We can confirm that EI, JES, and TS are inferior to others in both emulators.
In contrast, GP-UCB-based methods and PIMS show superior performance.
Thus, we can confirm the effectiveness of PIMS in both emulators derived from the real-world dataset without heuristic tunings, unlike GP-UCB-based methods.

\section{Conclusion}
\label{sec:conclusion}

First, we derived the tighter BCR bounds of TS.
Furthermore, we analyze the randomized variant of PI called PIMS, which alleviates the practical issues of GP-UCB-based methods and TS.
We showed the BCR bounds of PIMS, whose rate is the same as TS and IRGP-UCB.
Finally, we demonstrated the effectiveness of PIMS, particularly focusing on the comparison of PIMS, TS, and GP-UCB-based methods.

\paragraph{Limitations and Future Works}
%
Although the hyperparameters of AF do not exist, the hyperparameter tuning of GP is crucial.
Furthermore, the posterior sampling must be approximated by RFF for continuous $\cX$.
Therefore, regret analysis incorporating the uncertainty of GP hyperparameters \citep{Berkenkamp2019-no-regret,bogunovic2021misspecified} and the RFF approximation error \citep{Mutny2018-efficient} are important future directions.
%
%
%
Moreover, as discussed in Sec.~\ref{sec:related}, the regret analysis for the RKHS setting is also intriguing.
%
%
On the other hand, various extensions for, e.g., multi-fidelity \citep{Kandasamy2016-Gaussian}, constrained \citep{Gardner2014-Bayesian}, and multi-objective \citep{paria2020-flexible} optimization, are practically important.
In addition, EI based on the posterior sampling may be motivated as with PIMS.

\section*{Acknowkedgements}
This work was partially supported by 
MEXT Program: Data Creation and Utilization-Type Material Research and Development Project Grant Number JPMXP1122712807,
JSPS KAKENHI Grant Numbers JP20H00601, JP21H03498, JP22H00300, JP23K16943, JP23K19967, JP23K21696,
JST CREST Grant Numbers JPMJCR21D3, JPMJCR22N2, 
JST Moonshot R\&D Grant Number JPMJMS2033-05, 
JST AIP Acceleration Research Grant Number JPMJCR21U2, 
JST ACT-X Grant Number JPMJAX23CD,
NEDO Grant Numbers JPNP18002, JPNP20006, 
and RIKEN Center for Advanced Intelligence Project.

\bibliography{ref}

\begin{thebibliography}{40}
\providecommand{\natexlab}[1]{#1}
\providecommand{\url}[1]{\texttt{#1}}
\expandafter\ifx\csname urlstyle\endcsname\relax
  \providecommand{\doi}[1]{doi: #1}\else
  \providecommand{\doi}{doi: \begingroup \urlstyle{rm}\Url}\fi

\bibitem[Berkenkamp et~al.(2019)Berkenkamp, Schoellig, and Krause]{Berkenkamp2019-no-regret}
F.~Berkenkamp, A.~P. Schoellig, and A.~Krause.
\newblock No-regret {B}ayesian optimization with unknown hyperparameters.
\newblock \emph{Journal of Machine Learning Research}, 20\penalty0 (50):\penalty0 1--24, 2019.

\bibitem[Bogunovic and Krause(2021)]{bogunovic2021misspecified}
I.~Bogunovic and A.~Krause.
\newblock Misspecified {G}aussian process bandit optimization.
\newblock \emph{Advances in Neural Information Processing Systems 34}, pages 3004--3015, 2021.

\bibitem[Bull(2011)]{bull2011convergence}
A.~D. Bull.
\newblock Convergence rates of efficient global optimization algorithms.
\newblock \emph{Journal of Machine Learning Research}, 12\penalty0 (10), 2011.

\bibitem[Chowdhury and Gopalan(2017)]{Chowdhury2017-on}
S.~R. Chowdhury and A.~Gopalan.
\newblock On kernelized multi-armed bandits.
\newblock In \emph{Proceedings of the 34th International Conference on Machine Learning}, volume~70 of \emph{Proceedings of Machine Learning Research}, pages 844--853, 2017.

\bibitem[Falkner et~al.(2018)Falkner, Klein, and Hutter]{Falkner2019-BOHB}
S.~Falkner, A.~Klein, and F.~Hutter.
\newblock {BOHB}: Robust and efficient hyperparameter optimization at scale.
\newblock In \emph{Proceedings of the 35th International Conference on Machine Learning}, volume~80, pages 1437--1446. PMLR, 2018.

\bibitem[Frazier et~al.(2009)Frazier, Powell, and Dayanik]{Frazier2009-knowledge}
P.~Frazier, W.~Powell, and S.~Dayanik.
\newblock The knowledge-gradient policy for correlated normal beliefs.
\newblock \emph{INFORMS journal on Computing}, 21\penalty0 (4):\penalty0 599--613, 2009.

\bibitem[Gardner et~al.(2014)Gardner, Kusner, Zhixiang, Weinberger, and Cunningham]{Gardner2014-Bayesian}
J.~Gardner, M.~Kusner, Zhixiang, K.~Weinberger, and J.~Cunningham.
\newblock Bayesian optimization with inequality constraints.
\newblock In \emph{Proceedings of the 31st International Conference on Machine Learning}, volume~32, pages 937--945. PMLR, 2014.

\bibitem[H{\"a}se et~al.(2021)H{\"a}se, Aldeghi, Hickman, Roch, Christensen, Liles, Hein, and Aspuru-Guzik]{hase2021olympus}
F.~H{\"a}se, M.~Aldeghi, R.~J. Hickman, L.~M. Roch, M.~Christensen, E.~Liles, J.~E. Hein, and A.~Aspuru-Guzik.
\newblock Olympus: a benchmarking framework for noisy optimization and experiment planning.
\newblock \emph{Machine Learning: Science and Technology}, 2\penalty0 (3):\penalty0 035021, 2021.

\bibitem[Hennig and Schuler(2012)]{Henning2012-Entropy}
P.~Hennig and C.~J. Schuler.
\newblock Entropy search for information-efficient global optimization.
\newblock \emph{Journal of Machine Learning Research}, 13\penalty0 (57):\penalty0 1809--1837, 2012.

\bibitem[Hern\'{a}ndez-Lobato et~al.(2014)Hern\'{a}ndez-Lobato, Hoffman, and Ghahramani]{Hernandez2014-Predictive}
J.~M. Hern\'{a}ndez-Lobato, M.~W. Hoffman, and Z.~Ghahramani.
\newblock Predictive entropy search for efficient global optimization of black-box functions.
\newblock In \emph{Advances in Neural Information Processing Systems 27}, page 918–926. Curran Associates, Inc., 2014.

\bibitem[Hvarfner et~al.(2022)Hvarfner, Hutter, and Nardi]{hvarfner2022-joint}
C.~Hvarfner, F.~Hutter, and L.~Nardi.
\newblock Joint entropy search for maximally-informed {B}ayesian optimization.
\newblock In \emph{Advances in Neural Information Processing Systems 35}, pages 11494--11506. Curran Associates, Inc., 2022.

\bibitem[Janz et~al.(2020)Janz, Burt, and Gonzalez]{janz2020-bandit}
D.~Janz, D.~Burt, and J.~Gonzalez.
\newblock Bandit optimisation of functions in the {Matérn kernel RKHS}.
\newblock In \emph{Proceedings of the 23rd International Conference on Artificial Intelligence and Statistics}, volume 108 of \emph{Proceedings of Machine Learning Research}, pages 2486--2495, 2020.

\bibitem[Kandasamy et~al.(2015)Kandasamy, Schneider, and Poczos]{kandasamy2015-high}
K.~Kandasamy, J.~Schneider, and B.~Poczos.
\newblock High dimensional bayesian optimisation and bandits via additive models.
\newblock In \emph{Proceedings of the 32nd International Conference on Machine Learning}, volume~37 of \emph{Proceedings of Machine Learning Research}, pages 295--304, 2015.

\bibitem[Kandasamy et~al.(2016)Kandasamy, Dasarathy, Oliva, Schneider, and P{\'o}czos]{Kandasamy2016-Gaussian}
K.~Kandasamy, G.~Dasarathy, J.~Oliva, J.~Schneider, and B.~P{\'o}czos.
\newblock Gaussian process bandit optimisation with multi-fidelity evaluations.
\newblock In \emph{Advances in Neural Information Processing Systems 29}, pages 1000--1008. Curran Associates, Inc., 2016.

\bibitem[Kandasamy et~al.(2018)Kandasamy, Krishnamurthy, Schneider, and P{\'o}czos]{Kandasamy2018-Parallelised}
K.~Kandasamy, A.~Krishnamurthy, J.~Schneider, and B.~P{\'o}czos.
\newblock Parallelised {B}ayesian optimisation via {T}hompson sampling.
\newblock In \emph{Proceedings of the 21st International Conference on Artificial Intelligence and Statistics}, volume~84 of \emph{Proceedings of Machine Learning Research}, pages 133--142, 2018.

\bibitem[Klein et~al.(2017)Klein, Falkner, Bartels, Hennig, and Hutter]{Klein2017-Fast}
A.~Klein, S.~Falkner, S.~Bartels, P.~Hennig, and F.~Hutter.
\newblock Fast {B}ayesian optimization of machine learning hyperparameters on large datasets.
\newblock In \emph{Proceedings of the 20th International Conference on Artificial Intelligence and Statistics}, volume~54, pages 528--536. PMLR, 2017.

\bibitem[Korovina et~al.(2020)Korovina, Xu, Kandasamy, Neiswanger, Poczos, Schneider, and Xing]{korovina2020-chemBO}
K.~Korovina, S.~Xu, K.~Kandasamy, W.~Neiswanger, B.~Poczos, J.~Schneider, and E.~Xing.
\newblock Chembo: Bayesian optimization of small organic molecules with synthesizable recommendations.
\newblock In \emph{Proceedings of the 23rd International Conference on Artificial Intelligence and Statistics}, volume 108 of \emph{Proceedings of Machine Learning Research}, pages 3393--3403, 2020.

\bibitem[Kusakawa et~al.(2022)Kusakawa, Takeno, Inatsu, Kutsukake, Iwazaki, Nakano, Ujihara, Karasuyama, and Takeuchi]{Kusakawa2022-bayesian}
S.~Kusakawa, S.~Takeno, Y.~Inatsu, K.~Kutsukake, S.~Iwazaki, T.~Nakano, T.~Ujihara, M.~Karasuyama, and I.~Takeuchi.
\newblock {B}ayesian optimization for cascade-type multistage processes.
\newblock \emph{Neural Computation}, 34\penalty0 (12):\penalty0 2408--2431, 2022.

\bibitem[Kushner(1962)]{KUSHNER1962150}
H.~J. Kushner.
\newblock A versatile stochastic model of a function of unknown and time varying form.
\newblock \emph{Journal of Mathematical Analysis and Applications}, 5\penalty0 (1):\penalty0 150--167, 1962.

\bibitem[Kushner(1964)]{Kushner1964-new}
H.~J. Kushner.
\newblock {A New Method of Locating the Maximum Point of an Arbitrary Multipeak Curve in the Presence of Noise}.
\newblock \emph{Journal of Basic Engineering}, 86\penalty0 (1):\penalty0 97--106, 1964.

\bibitem[Loh(1996)]{Loh1996-Latin}
W.-L. Loh.
\newblock On {L}atin hypercube sampling.
\newblock \emph{Ann. Statist.}, 24\penalty0 (5):\penalty0 2058--2080, 1996.

\bibitem[Mockus et~al.(1978)Mockus, Tiesis, and Zilinskas]{Mockus1978-Application}
J.~Mockus, V.~Tiesis, and A.~Zilinskas.
\newblock The application of {B}ayesian methods for seeking the extremum.
\newblock \emph{Towards Global Optimization}, 2\penalty0 (117-129):\penalty0 2, 1978.

\bibitem[Mutny and Krause(2018)]{Mutny2018-efficient}
M.~Mutny and A.~Krause.
\newblock Efficient high dimensional {B}ayesian optimization with additivity and quadrature {F}ourier features.
\newblock \emph{Advances in Neural Information Processing Systems 31}, pages 9005--9016, 2018.

\bibitem[Paria et~al.(2020)Paria, Kandasamy, and P{\'{o}}czos]{paria2020-flexible}
B.~Paria, K.~Kandasamy, and B.~P{\'{o}}czos.
\newblock A flexible framework for multi-objective {B}ayesian optimization using random scalarizations.
\newblock In \emph{Proceedings of The 35th Uncertainty in Artificial Intelligence Conference}, volume 115 of \emph{Proceedings of Machine Learning Research}, pages 766--776, 2020.

\bibitem[Rahimi and Recht(2008)]{Rahimi2008-Random}
A.~Rahimi and B.~Recht.
\newblock Random features for large-scale kernel machines.
\newblock In \emph{Advances in Neural Information Processing Systems 20}, pages 1177--1184. Curran Associates, Inc., 2008.

\bibitem[Rasmussen and Williams(2005)]{Rasmussen2005-Gaussian}
C.~E. Rasmussen and C.~K.~I. Williams.
\newblock \emph{Gaussian Processes for Machine Learning (Adaptive Computation and Machine Learning)}.
\newblock The MIT Press, 2005.

\bibitem[Russo and Van~Roy(2014)]{Russo2014-learning}
D.~Russo and B.~Van~Roy.
\newblock Learning to optimize via posterior sampling.
\newblock \emph{Mathematics of Operations Research}, 39\penalty0 (4):\penalty0 1221--1243, 2014.

\bibitem[Shahriari et~al.(2016)Shahriari, Swersky, Wang, Adams, and {De Freitas}]{Shahriari2016-Taking}
B.~Shahriari, K.~Swersky, Z.~Wang, R.~Adams, and N.~{De Freitas}.
\newblock Taking the human out of the loop: A review of {B}ayesian optimization.
\newblock \emph{Proceedings of the IEEE}, 104\penalty0 (1):\penalty0 148--175, 2016.

\bibitem[Srinivas et~al.(2010)Srinivas, Krause, Kakade, and Seeger]{Srinivas2010-Gaussian}
N.~Srinivas, A.~Krause, S.~Kakade, and M.~Seeger.
\newblock Gaussian process optimization in the bandit setting: No regret and experimental design.
\newblock In \emph{Proceedings of the 27th International Conference on Machine Learning}, pages 1015--1022. Omnipress, 2010.

\bibitem[Takeno et~al.(2022)Takeno, Tamura, Shitara, and Karasuyama]{takeno2022-sequential}
S.~Takeno, T.~Tamura, K.~Shitara, and M.~Karasuyama.
\newblock Sequential and parallel constrained max-value entropy search via information lower bound.
\newblock In \emph{Proceedings of the 39th International Conference on Machine Learning}, volume 162 of \emph{Proceedings of Machine Learning Research}, pages 20960--20986, 2022.

\bibitem[Takeno et~al.(2023)Takeno, Inatsu, and Karasuyama]{Takeno2023-randomized}
S.~Takeno, Y.~Inatsu, and M.~Karasuyama.
\newblock Randomized {G}aussian process upper confidence bound with tighter {B}ayesian regret bounds.
\newblock In \emph{Proceedings of the 40th International Conference on Machine Learning}, volume 202 of \emph{Proceedings of Machine Learning Research}, pages 33490--33515. PMLR, 2023.

\bibitem[Thompson(1933)]{Thompson1933-likelihood}
W.~R. Thompson.
\newblock On the likelihood that one unknown probability exceeds another in view of the evidence of two samples.
\newblock \emph{Biometrika}, 25\penalty0 (3-4):\penalty0 285--294, 1933.

\bibitem[Ueno et~al.(2016)Ueno, Rhone, Hou, Mizoguchi, and Tsuda]{ueno2016combo}
T.~Ueno, T.~D. Rhone, Z.~Hou, T.~Mizoguchi, and K.~Tsuda.
\newblock {COMBO}: An efficient {B}ayesian optimization library for materials science.
\newblock \emph{Materials discovery}, 4:\penalty0 18--21, 2016.

\bibitem[Vakili et~al.(2021)Vakili, Khezeli, and Picheny]{vakili2021-information}
S.~Vakili, K.~Khezeli, and V.~Picheny.
\newblock On information gain and regret bounds in {G}aussian process bandits.
\newblock In \emph{Proceedings of The 24th International Conference on Artificial Intelligence and Statistics}, volume 130 of \emph{Proceedings of Machine Learning Research}, pages 82--90, 2021.

\bibitem[Villemonteix et~al.(2009)Villemonteix, Vazquez, and Walter]{Villemonteix2009-aninformational}
J.~Villemonteix, E.~Vazquez, and E.~Walter.
\newblock {An informational approach to the global optimization of expensive-to-evaluate functions}.
\newblock \emph{{Journal of Global Optimization}}, 44\penalty0 (4):\penalty0 509--534, 2009.

\bibitem[Walker et~al.(2017)Walker, Bannock, Nightingale, and deMello]{Walker2017-tuning}
B.~E. Walker, J.~H. Bannock, A.~M. Nightingale, and J.~C. deMello.
\newblock Tuning reaction products by constrained optimisation.
\newblock \emph{React. Chem. Eng.}, 2:\penalty0 785--798, 2017.

\bibitem[Wang and de~Freitas(2014)]{wang2014theoretical}
Z.~Wang and N.~de~Freitas.
\newblock Theoretical analysis of {B}ayesian optimisation with unknown {G}aussian process hyper-parameters.
\newblock \emph{arXiv:1406.7758}, 2014.

\bibitem[Wang and Jegelka(2017)]{Wang2017-Max}
Z.~Wang and S.~Jegelka.
\newblock Max-value entropy search for efficient {B}ayesian optimization.
\newblock In \emph{Proceedings of the 34th International Conference on Machine Learning}, volume~70 of \emph{Proceedings of Machine Learning Research}, pages 3627--3635, 2017.

\bibitem[Wang et~al.(2016)Wang, Zhou, and Jegelka]{Wang2016-Optimization}
Z.~Wang, B.~Zhou, and S.~Jegelka.
\newblock Optimization as estimation with {G}aussian processes in bandit settings.
\newblock In \emph{Proceedings of the 19th International Conference on Artificial Intelligence and Statistics}, volume~51 of \emph{Proceedings of Machine Learning Research}, pages 1022--1031, 2016.

\bibitem[Wang et~al.(2018)Wang, Kim, and Kaelbling]{Wang2018-regret}
Z.~Wang, B.~Kim, and L.~P. Kaelbling.
\newblock Regret bounds for meta {B}ayesian optimization with an unknown {G}aussian process prior.
\newblock In \emph{Advances in Neural Information Processing Systems 31}, pages 10477--10488. Curran Associates, Inc., 2018.

\end{thebibliography}
\bibliographystyle{abbrvnat}

\clearpage
\appendix
\onecolumn

\section{Equivalence between PIMS and Existing Methods}
\label{app:equivalence}

\citet{Wang2016-Optimization, Wang2017-Max} claimed the following lemma:
\begin{lemma}
    The following methods are equivalent:
    \vspace{-1em}
    \begin{enumerate}
        \item PIMS
        \item MES with only one MC sample
        \item (randomized) GP-UCB that maximize $\mu_{t-1}(\*x) + \xi_t \sigma_{t-1}(\*x)$, where $\xi_t = \min_{\*x \in \cX} \left\{ \bigl( g^*_t - \mu_{t-1}(\*x) \bigr) / \sigma_{t-1}(\*x) \right\}$.
        \item (GP-EST using $g^*_t$ instead of $\hat{m}$ defined in \citet{Wang2016-Optimization})
    \end{enumerate}
    \label{lem:equivalence_PIMS}
\end{lemma}
First, the equivalence between PIMS and MES with only one MC sample is shown in Lemma~3.1 in \citet{Wang2017-Max}.
This equivalence is obvious from the monotonicity of MES's AF when the MC sample size is one.
Then, for the equivalence between PIMS and GP-UCB, \citet{Wang2016-Optimization} provided the proof in Lemma~2.1.
However, since this proof is slightly ambiguous, we provide the proof for completeness.

Let $\*x_{\rm UCB}$ and $\*x_{\rm PIMS}$ be the selected points by GP-UCB and PIMS, i.e.,
\begin{align*}
    \*x_{\rm UCB} &\in \argmax_{\*x \in \cX} \left\{ \mu_{t-1}(\*x) + \xi_t \sigma_{t-1}(\*x) \right\}, \\
    \*x_{\rm PIMS} &\in \argmin_{\*x \in \cX} \left\{ \frac{g^*_t - \mu_{t-1}(\*x)}{\sigma_{t-1}(\*x)} \right\}.
\end{align*}
Fix $g^*_t \in \RR$.
From the definitions of GP-UCB, PIMS, and $\xi_t$, the following holds:
\begin{align*}
    \mu_{t-1}(\*x_{\rm UCB}) + \xi_t \sigma_{t-1}(\*x_{\rm UCB}) 
    &\geq \mu_{t-1}(\*x_{\rm PIMS}) + \xi_t \sigma_{t-1}(\*x_{\rm PIMS}) \\
    &= \mu_{t-1}(\*x_{\rm PIMS}) +  \frac{g^*_t - \mu_{t-1}(\*x_{\rm PIMS})}{\sigma_{t-1}(\*x_{\rm PIMS})} \sigma_{t-1}(\*x_{\rm PIMS}) \\
    &= g^*_t.
\end{align*}
Then, if $\mu_{t-1}(\*x_{\rm UCB}) + \xi_t \sigma_{t-1}(\*x_{\rm UCB}) > g^*_t$, we can see that
\begin{align*}
    \xi_t > \frac{g^*_t - \mu_{t-1}(\*x_{\rm UCB})}{\sigma_{t-1}(\*x_{\rm UCB})}.
\end{align*}
This clearly contradicts the definition of $\xi_t$.
Therefore, 
\begin{align*}
    \mu_{t-1}(\*x_{\rm UCB}) + \xi_t \sigma_{t-1}(\*x_{\rm UCB}) 
    &= \mu_{t-1}(\*x_{\rm PIMS}) + \xi_t \sigma_{t-1}(\*x_{\rm PIMS}) \\
    &= g^*_t.
\end{align*}
In addition, we can transform the inequality as follows:
\begin{align*}
    \frac{g^*_t - \mu_{t-1}(\*x_{\rm UCB})}{\sigma_{t-1}(\*x_{\rm UCB})}
    &= \frac{g^*_t - \mu_{t-1}(\*x_{\rm PIMS})}{\sigma_{t-1}(\*x_{\rm PIMS})} \\
    &= \xi_t.
\end{align*}
Consequently, since the above equalities hold for all $g^*_t \in \RR$, we can obtain 
\begin{align*}
    \*x_{\rm UCB}, \*x_{\rm PIMS} \in \argmax_{\*x \in \cX} \left\{ \mu_{t-1}(\*x) + \xi_t \sigma_{t-1}(\*x) \right\}, \\
    \*x_{\rm UCB}, \*x_{\rm PIMS} \in \argmin_{\*x \in \cX} \left\{ \frac{g^*_t - \mu_{t-1}(\*x)}{\sigma_{t-1}(\*x)} \right\}.
\end{align*}

\section{Proof regarding TS}
\label{app:TS}

First, we show the proof of the following lemma:
\begin{replemma}{lem:bound_TS}
    Let $f \sim \cG \cP (0, k)$, where $k$ is a stationary kernel and $k(\*x, \*x) = 1$, and $\cX$ be finite.
    Let $\eta_t \coloneqq \frac{g_t(\*x_t) - \mu_{t-1}(\*x_t)}{\sigma_{t-1}(\*x_t)}$, where $g_t \sim p(f | \cD_{t-1})$ and $\*x_t = \argmax_{\*x \in \cX} g_t(\*x)$.
    Then, for all $t \geq 1$, the following inequality holds:
    \begin{align*}
        \EE \left[ \eta^2_t \mathbbm{1} \{ \eta_t \geq 0\} \right] 
        \leq 2 + 2 \log \bigl(|\cX|/2\bigr),
    \end{align*}
    where $\mathbbm{1}\{ \eta_t \geq 0 \} = 1$ if $\eta_t \geq 0$, and otherwise $0$.
\end{replemma}
\begin{proof}
    It suffices to show that, for an arbitrary $\cD_{t-1}$, 
    \begin{align*}
        \EE_t \left[ \eta^2_t \mathbbm{1} \{ \eta_t \geq 0\} \right] 
        \leq 2 + 2 \log \bigl(|\cX|/2\bigr).
    \end{align*}
    %
    %
    Then, fix the dataset $\cD_{t-1}$ and $\delta \in (0, 1)$.
    %
    %
    From Lemma~\ref{lem:bound_srinivas}, we can see that
    \begin{align*}
        \myPr_t \left( f(\*x^*) \leq \mu_{t-1}(\*x^*) + \beta^{1/2}_{\delta} \sigma_{t-1}(\*x^*) \right)
        &\geq
        \myPr_t \left( \forall \*x \in \cX,\ f(\*x) \leq \mu_{t-1}(\*x) + \beta^{1/2}_{\delta} \sigma_{t-1}(\*x) \right) \\
        &\geq 1 - \delta.
    \end{align*}
    where $\beta_{\delta} =  2 \log (|\cX| / (2\delta))$.
    Then, since $g_t, \*x_t \mid \cD_{t-1} \disteq f, \*x^* \mid \cD_{t-1}$, where $\disteq$ implies the equality in distribution, we can arrange the probability as follows:
    \begin{align*}
        \myPr_t \left( f(\*x^*) \leq \mu_{t-1}(\*x^*) + \beta^{1/2}_{\delta} \sigma_{t-1}(\*x^*) \right)
        &=
         \myPr_t \left( \frac{f(\*x^*) - \mu_{t-1}(\*x^*)}{\sigma_{t-1}(\*x^*)} \leq \beta^{1/2}_{\delta} \right) \mybecause{\sigma_{t-1}(\*x^*) > 0} \\
         &= \myPr_t \left(  \eta_t \leq \beta^{1/2}_{\delta} \right) \mybecause{g_t, \*x_t \mid \cD_{t-1} \disteq f, \*x^* \mid \cD_{t-1}} \\
         &= \myPr_t \left(  \eta_t \mathbbm{1} \{ \eta_t \geq 0\} \leq \beta^{1/2}_{\delta} \right) \mybecause{\beta^{1/2}_{\delta} > 0}.
    \end{align*}
    Therefore, we obtain
    \begin{align*}
        \myPr_t \left( \eta^2_t \mathbbm{1} \{ \eta_t \geq 0\} \leq \beta_{\delta} \right) 
        \geq 1 - \delta.
    \end{align*}
    Using the cumulative distribution function $F_t(\cdot) \coloneqq \myPr_t( \eta^2_t \mathbbm{1} \{ \eta_t \geq 0\} \leq \cdot)$ and its generalized inverse function $F^{-1}_t$ (``generalized'' is required since $\eta^2_t \mathbbm{1} \{ \eta_t \geq 0\}$ is not an absolute continuous random variable), we can rewrite
    \begin{align*}
        F_t \left( \beta_{\delta} \right) 
        \geq 1 - \delta
        \Longleftrightarrow 
        \beta_{\delta} 
        \geq F^{-1}_t ( 1 - \delta ),
    \end{align*}
    since $F^{-1}_t$ is a monotone non-decreasing function.
    Since $\delta \in (0, 1)$ is arbitrary, we can substitute $U \sim {\rm Uni}(0, 1)$ as,
    \begin{align*}
         \beta_{U} \geq F^{-1}_t ( 1 - U ).
    \end{align*}
    By taking the expectation with respect to $U$, we can obtain,
    \begin{align*}
         \EE_{U} \left[ \beta_{U} \right]
         &\geq \EE_{U} \bigl[ F^{-1}_t ( 1 - U )   \bigr] \\
         &= \EE_{U} \bigl[ F^{-1}_t ( U )   \bigr],
    \end{align*}
    where we use the fact that $1 - U$ also follows ${\rm Uni}(0, 1)$.
    Since $F^{-1}_t ( U )$ and $\eta^2_t \mathbbm{1} \{ \eta_t \geq 0\} | \cD_{t-1}$ are identically distributed as with the inverse transform sampling, we obtain
    \begin{align*}
        \EE_t \left[ \eta^2_t \mathbbm{1} \{ \eta_t \geq 0\} \right] 
        \leq \EE_{U} \left[ \beta_{U} \right]
        = 2 + 2 \log \bigl(|\cX|/2\bigr),
    \end{align*}
    which concludes the proof.
\end{proof}

\subsection{Discrete Domain}
\label{app:TS_discrete}

\begin{reptheorem}{theo:BCR_TS_discrete}
    Let $f \sim \cG \cP (0, k)$, where $k$ is a stationary kernel and $k(\*x, \*x) = 1$, and $\cX$ be finite.
    Then, by running TS, BCR can be bounded as follows: 
    \begin{align*}
        {\rm BCR}_T \leq \sqrt{C_1 C_2 T \gamma_T},
    \end{align*}
    where $C_1 \coloneqq 2 / \log(1 + \sigma^{-2})$ and $C_2 \coloneqq 2 + 2 \log \bigl(|\cX|/2\bigr)$. 
\end{reptheorem}
\begin{proof}
    We can obtain the following upper bound:
    \begin{align*}
        {\rm BCR}_T 
        &= \sum_{t=1}^T \EE\left[ f(\*x^*) - g_t(\*x_t) + g_t(\*x_t) - f(\*x_t) \right] \\
        &= \sum_{t=1}^T \EE_{\cD_{t-1}}\left[ \EE_t \left[ f(\*x^*) - g_t(\*x_t) \right] \right] + \sum_{t=1}^T \EE\left[g_t(\*x_t) - f(\*x_t) \right] \\
        &= \sum_{t=1}^T \EE\left[g_t(\*x_t) - f(\*x_t) \right] \\
        &= \sum_{t=1}^T \EE_{\cD_{t-1}}\left[ \EE_t \left[ g_t(\*x_t) - f(\*x_t) \right] \right] \\
        &= \sum_{t=1}^T \EE \left[ g_t(\*x_t) - \mu_{t-1}(\*x_t) \right].
    \end{align*}
    Then, we apply Lemma~\ref{lem:bound_TS} to $\eta_t \coloneqq \frac{g_t(\*x_t) - \mu_{t-1}(\*x_t)}{\sigma_{t-1}(\*x_t)}$.
    Therefore, we obtain
    \begin{align*}
        {\rm BCR}_T 
        &= \sum_{t=1}^T \EE \left[ g_t(\*x_t) - \mu_{t-1}(\*x_t) \right] \\
        &= \sum_{t=1}^T \EE \left[ \eta_t \sigma_{t-1}(\*x_t) \right] \\
        &\leq \sum_{t=1}^T \EE \left[ \eta_t \mathbbm{1} \{ \eta_t \geq 0\} \sigma_{t-1}(\*x_t) \right] \\
        &\leq \EE \left[ \sqrt{ \sum_{t=1}^T \eta^2_t \mathbbm{1} \{ \eta_t \geq 0\} \sum_{t=1}^T \sigma^2_{t-1}(\*x_t)} \right] && \bigl(\because \text{Cauchy--Schwarz inequality} \bigr) \\
        &\leq \EE \left[ \sqrt{ \sum_{t=1}^T \eta^2_t \mathbbm{1} \{ \eta_t \geq 0\} } \right] \sqrt{C_1 \gamma_T} && \bigl(\because \text{Lemma~5.4 in \citep{Srinivas2010-Gaussian}} \bigr) \\
        &\leq \sqrt{ \EE \left[ \sum_{t=1}^T \eta^2_t \mathbbm{1} \{ \eta_t \geq 0\} \right]} \sqrt{C_1 \gamma_T} && \bigl(\because \text{Jensen inequality} \bigr) \\
        &\leq\sqrt{C_1 C_2 T\gamma_T} && \bigl(\because \text{Lemma~\ref{lem:bound_TS}} \bigr),
    \end{align*}
    where $C_2 \coloneqq 2 + 2 \log \bigl(|\cX|/2\bigr)$.
\end{proof}

\subsection{Continuous Domain with Continuous Sample Path}
\label{app:TS_continuous}

First, we provide the rigorous version of Lemma~\ref{lem:Lipschitz_std_short}:
\begin{lemma}[Theorem~E.4 in \citep{Kusakawa2022-bayesian}, Lipschitz constants for posterior standard deviation]
    Let $k(\*x, \*x^\prime): \RR^d \times \RR^d \to \RR$ be one of the following kernel functions: 
    \begin{description}
    \item [Linear kernel:] $k(\*x,\*x^\prime) = \sigma_{f}^2 \*x^\top \*x^\prime$, where $\sigma_{f}$ is a positive parameter. 
    \item [Gaussian kernel:] $k(\*x,\*x^\prime) = \sigma_{f}^2 \exp (-\| \*x - \*x^\prime \|^2/(2 \ell^2 ) )$, where $\sigma_{f}$ and $ \ell$ are positive parameters. 
    \item [Mat\'{e}rn kernel:] 
    $$
        k(\*x,\*x^\prime) = \sigma_{f}^2 \frac{ 2^{1-\nu} }{\Gamma (\nu)}  \left ( \sqrt{2 \nu} \frac{\| \*x - \*x^\prime \|}{\ell} \right )^\nu K_\nu \left (   \sqrt{2 \nu}   \frac{ \| \*x - \*x^\prime \| }{\ell} \right ),
    $$
    where $\sigma_{f}$ and $\ell$ are positive parameters, $\nu$ is a degree of freedom with $\nu >1$, $\Gamma$ is the gamma function, and $K_\nu$ is the modified Bessel function of the second kind.
    \end{description}
    Moreover, assume that a noise variance $\sigma^2$ is positive.
    Then, for any $t \geq 1$ and $\cD_{t-1}$, the posterior standard deviation $\sigma_{t-1} (\*x )$ satisfies that 
    \begin{align*}
        \forall \*x,\*x^\prime \in \RR^d, \ | \sigma_{t-1} (\*x ) - \sigma_{t-1} (\*x^\prime ) | \leq L_{\sigma} \| \*x - \*x^\prime \| _1,
    \end{align*}
    where $L_{\sigma}$ is a positive constant given by 
    \begin{align*}
        L_{\sigma} = \left\{
        \begin{array}{ll}
            \sigma_{f} & \text{if $k(\*x,\*x^\prime)$ is the linear kernel}, \\
            \frac{\sqrt{2} \sigma_{f}}{\ell} & \text{if $k(\*x,\*x^\prime)$ is the Gaussian kernel}, \\
            \frac{\sqrt{2} \sigma_{f}}{\ell}
            \sqrt{\frac{\nu}{\nu-1} } & \text{if $k(\*x,\*x^\prime)$ is the Mat\'{e}rn kernel},
        \end{array}\right.
    \end{align*}
    \label{lem:Lipschitz_std}
\end{lemma}

Next, we show the following BCR bounds of TS:
\begin{reptheorem}{theo:BCR_TS_continuous}
    Let $f \sim \cG \cP (0, k)$, where $k$ is a stationary kernel, $k(\*x, \*x) = 1$, and Assumption~\ref{assump:continuous_X} holds.
    Let $L = \max \bigl\{L_\sigma, b \bigl(\sqrt{\log (ad)} + \sqrt{\pi} / 2 \bigr) \bigr\}$, where $L_\sigma$ is defined as in Lemma~\ref{lem:Lipschitz_std_short}.
    Then, by running TS, BCR can be bounded as follows: 
    \begin{align*}
        {\rm BCR}_T \leq \frac{\pi^2}{3} + \frac{\pi^2}{6} \sqrt{s_T} + \sqrt{ C_1 \gamma_T T s_T},
    \end{align*}
    where $C_1 \coloneqq 2 / \log(1 + \sigma^{-2})$ and $s_t = 2 - 2 \log 2 + 2d \log(\lceil drL t^2 \rceil)$. 
\end{reptheorem}
\begin{proof}
    For the sake of analysis, we used a set of discretization $\cX_t \subset \cX$ for $t \geq 1$.
    For any $t \geq 1$, let $\cX_t \subset \cX$ be a finite set with each dimension equally divided into $\tau_t = \lceil drL t^2 \rceil$.
    Thus, $|\cX_t| = \tau_t^d$.
    In addition, we define $[\*x]_t$ as the nearest point in $\cX_t$ of $\*x \in \cX$.

    Then, we decompose BCR as follows:
    \begin{align*}
        {\rm BCR}_T
        &= \sum_{t=1}^T \EE \Biggl[ 
            f(\*x^*) - f( [\*x^*]_t )
            + f( [\*x^*]_t ) - g_t( [\*x_t]_t )
            + g_t( [\*x_t]_t ) - f([\*x_t]_t)
            + f([\*x_t]_t) - f(\*x_t)
            \Biggl] \\
        &= \underbrace{\EE \Biggl[ \sum_{t=1}^T f(\*x^*) - f( [\*x^*]_t ) \Biggl]}_{A_1}
        + \underbrace{\EE \Biggl[ \sum_{t=1}^T f( [\*x^*]_t ) - g_t( [\*x_t]_t ) \Biggl]}_{A_2} 
        \\ & \qquad 
        + \underbrace{\EE \Biggl[ \sum_{t=1}^T g_t( [\*x_t]_t ) - f([\*x_t]_t) \Biggl]}_{A_3}
        + \underbrace{\EE \Biggl[ \sum_{t=1}^T f([\*x_t]_t) - f(\*x_t) \Biggl]}_{A_4}.
    \end{align*}
    Then, using Lemma~\ref{lem:discretized_error_sample_path}, terms $A_1$ and $A_4$ can be bounded above by $\pi^2 / 6$, respectively.
    Furthermore, from the definition, $g_t([\*x_t]_t) | \cD_{t-1}$ and $f([\*x^*]_t) | \cD_{t-1}$ are identically distributed.
    Hence, we can see that 
    \begin{align*}
        A_2
        = \sum_{t=1}^T \EE_{\cD_{t-1}} \bigl[ \EE_t \bigl[ f( [\*x^*]_t ) - g_t( [\*x_t]_t ) \bigr] \bigr]
        = 0.
    \end{align*}

    Finally, we need to bound $A_3$.
    As with Lemma~\ref{lem:bound_TS}, for all $t \geq 0$, we can obtain
    \begin{align*}
        \EE \left[ \eta^2_t \mathbbm{1} \{ \eta_t \geq 0\} \right] 
        \leq 2 + 2 \log \bigl(|\cX_t|/2\bigr),
    \end{align*}
    where $\eta_t \coloneqq \frac{g_t( [\*x_t]_t ) - \mu_{t-1}( [\*x_t]_t )}{\sigma_{t-1}( [\*x_t]_t )} $.
    Note that the proof for Lemma~\ref{lem:bound_TS} can apply to any distribution over $\cX_t$ including $\myPr_t([\*x_t]_t)$, not only $\myPr_t(\*x_t)$.
    Therefore, we can obtain 
    \begin{align*}
        A_3
        &= \EE \left[ \sum_{t=1}^T \eta_t \sigma_{t-1}([\*x_t]_t) \right]\\
        &\leq \EE \left[ \sum_{t=1}^T \eta_t \mathbbm{1}\{\eta_t \geq 0\} \sigma_{t-1}([\*x_t]_t) \right] \\
        &\leq 
            \underbrace{\EE \left[ \sum_{t=1}^T \eta_t \mathbbm{1}\{\eta_t \geq 0\} \sigma_{t-1}(\*x_t) \right]}_{A_5}
            + 
            \underbrace{\EE \left[ \sum_{t=1}^T \eta_t \mathbbm{1}\{\eta_t \geq 0\} |\sigma_{t-1}([\*x_t]_t) - \sigma_{t-1}(\*x_t) | \right]}_{A_6}.
    \end{align*}
    Then, as with Theorem~\ref{theo:BCR_TS_discrete}, we can see that
    \begin{align*}
        A_5 \leq \EE \left[ \sqrt{ \sum_{t=1}^T \eta^2_t \mathbbm{1}\{\eta_t \geq 0\} \sum_{t=1}^T \sigma^2_{t-1}(\*x_t )} \right]
        \leq \sqrt{C_1 \gamma_T T s_T},
    \end{align*}
    where $s_T = 2 + 2 \log\left( |\cX_t| / 2 \right) = 2 - \log 2 + 2d \log(\lceil drL t^2 \rceil)$.
    Furthermore, we can obtain
    \begin{align*}
        A_6 
        &\leq \EE \left[ \sum_{t=1}^T \eta_t \mathbbm{1}\{\eta_t \geq 0\} \frac{1}{t^2} \right] \mybecause{\text{Lemma~\ref{lem:discretized_error_std} based on Lemma~\ref{lem:Lipschitz_std}}} \\
        &= \sum_{t=1}^T \EE [ \eta_t \mathbbm{1}\{\eta_t \geq 0\} ] \frac{1}{t^2} \\
        &\leq \sum_{t=1}^T \sqrt{\EE [ \eta^2_t \mathbbm{1}\{\eta_t \geq 0\} ] }\frac{1}{t^2} \mybecause{\text{Jensen's inequality}} \\
        &\leq \sum_{t=1}^T \sqrt{2 + 2\log\bigl(|\cX_t|/2\bigr) }\frac{1}{t^2} \\
        &\leq \sqrt{2 + 2\log\bigl(|\cX_T|/2\bigr) } \sum_{t=1}^T\frac{1}{t^2} \mybecause{\text{$|\cX_t|$ is monotonically increasing}} \\
        &\leq \frac{\pi^2}{6} \sqrt{s_T}
    \end{align*}
    which concludes the proof.
\end{proof}

\subsection{Continuous Domain with Discretized Sample Path}
\label{app:disretizedTS_continuous}

If we modify TS so that $[\*x_t]_t$ is evaluated, the following theorem holds:
\begin{theorem}
    Let $f \sim \cG \cP (0, k)$, where $k$ is a stationary kernel, $k(\*x, \*x) = 1$, and Assumption~\ref{assump:continuous_X} holds.
    Let $L = b \bigl(\sqrt{\log (ad)} + \sqrt{\pi} / 2 \bigr)$.
    Then, by running modified TS that evaluates $[\*x_t]_t$, BCR can be bounded as follows: 
    \begin{align*}
        {\rm BCR}_T \leq \frac{\pi^2}{6} + \sqrt{ C_1 \gamma_T T s_T},
    \end{align*}
    where $C_1 \coloneqq 2 / \log(1 + \sigma^{-2})$ and $s_t = 2 - 2 \log 2 + 2d \log(\lceil drL t^2 \rceil)$. 
\end{theorem}
\begin{proof}
    We can decompose BCR as follows:
    \begin{align*}
        {\rm BCR}_T
        &= \sum_{t=1}^T \EE \Biggl[ 
            f(\*x^*) - f( [\*x^*]_t )
            + f( [\*x^*]_t ) - g_t( [\*x_t]_t )
            + g_t( [\*x_t]_t ) - f([\*x_t]_t)
            \Biggl] \\
        &= \underbrace{\EE \Biggl[ \sum_{t=1}^T f(\*x^*) - f( [\*x^*]_t ) \Biggl]}_{A_1}
        + \underbrace{\EE \Biggl[ \sum_{t=1}^T f( [\*x^*]_t ) - g_t( [\*x_t]_t ) \Biggl]}_{A_2} 
         + \underbrace{\EE \Biggl[ \sum_{t=1}^T g_t( [\*x_t]_t ) - f([\*x_t]_t) \Biggl]}_{A_3}.
    \end{align*}
    Then, as with the proof of Theorem~\ref{theo:BCR_TS_continuous}, we can obtain $A_1 \leq \frac{\pi^2}{6}$ and $A_2 = 0$.
    Furthermore, we can obtain
    \begin{align*}
        A_3 \leq \EE \left[ \sum_{t=1}^T \eta_t \mathbbm{1}\{\eta_t \geq 0\} \sigma_{t-1}([\*x_t]_t) \right].
    \end{align*}
    Since modified TS evaluates $[\*x_t]_t$, we can directly bound $A_3$ using MIG and $L = b \bigl(\sqrt{\log (ad)} + \sqrt{\pi} / 2 \bigr)$, i.e., 
    \begin{align*}
        A_3 \leq \sqrt{C_1 \gamma_T T s_T}.
    \end{align*}
\end{proof}

\section{Proof regarding PIMS}
\label{app:PIMS}

First, we show the following lemma:
\begin{replemma}{lem:bound_PIMS}
    Let $f \sim \cG \cP (0, k)$, where $k$ is a stationary kernel and $k(\*x, \*x) = 1$, and $\cX$ be finite.
    Let $\xi_t \coloneqq \min_{\*x \in \cX} \frac{g^*_t - \mu_{t-1}(\*x)}{\sigma_{t-1}(\*x)}$.
    Then, the following inequality holds:
    \begin{align*}
        \EE \left[ \xi^2_t \mathbbm{1} \{ \xi_t \geq 0\} \right] 
        \leq 2 + 2 \log \bigl(|\cX|/2\bigr),
    \end{align*}
    for all $t \geq 1$.
\end{replemma}
\begin{proof}
    %
    From the property of non-negative random variable,
    \begin{align}
        \EE\Bigl[  \xi^2_{t}\mathbbm{1}\{ \xi_{t} \geq 0 \}  \Bigr]
        &= \int^{\infty}_{0} \Pr\Bigl( \xi^2_{t}\mathbbm{1}\{ \xi_{t} \geq 0 \} > c \Bigl) {\rm d}c.
        \label{eq:acq_square_bound}
    \end{align}
    Using Lemmas~\ref{lem:Gauss_tail_bound}, \ref{lem:acq_tail_bound}, and union bound, for all $\cD_{t-1}$ and $c \geq 0$,
    \begin{align*}
        \myPr_t \Bigl(\xi^2_{t} \mathbbm{1}\{ \xi_{t} \geq 0 \} > c\Bigl)
        &= \myPr_t \Bigl(\xi_{t} > \sqrt{c} \Bigl) \\
        &= \myPr_t \bigl( g^*_t > \max_{\*x \in \cX} \left\{ \mu_{t-1}(\*x) + \sqrt{c} \sigma_{t-1} (\*x) \right\} \bigr) \mybecause{\text{Lemma~\ref{lem:acq_tail_bound}}}\\
        &= \myPr_t \bigl( \exists \*x,  g_t(\*x) > \max_{\*x \in \cX} \left\{ \mu_{t-1}(\*x) + \sqrt{c} \sigma_{t-1} (\*x) \right\} \bigr) \\
        &= \myPr_t \bigl( \exists \*x,  f(\*x) > \max_{\*x \in \cX} \left\{ \mu_{t-1}(\*x) + \sqrt{c} \sigma_{t-1} (\*x) \right\} \bigr) \\
        &\leq \myPr_t \bigl( \exists \*x, f(\*x) > \mu_{t-1}(\*x) + \sqrt{c} \sigma_{t-1} (\*x)  \bigr) \\
        &\leq \sum_{\*x \in \cX} \myPr_t \bigl( f(\*x) > \mu_{t-1}(\*x) + \sqrt{c} \sigma_{t-1} (\*x)  \bigr) \mybecause{\text{union bound}}\\
        &\leq \frac{|\cX|}{2} e^{-c / 2}. \mybecause{\text{Lemma~\ref{lem:Gauss_tail_bound}}}
    \end{align*}
    Therefore, we obtain
    \begin{align*}
        \Pr ( \xi^2_t \mathbbm{1}\{ \xi_{t} \geq 0 \} > c) = \EE_{\cD_{t-1}} \bigl[ \myPr_t ( \xi^2_t \mathbbm{1}\{ \xi_{t} \geq 0 \} > c ) \bigr] \leq \frac{|\cX|}{2} e^{-c / 2}.
    \end{align*}
    Since $ \Pr\Bigl(\xi^2_{t} \mathbbm{1}\{ \xi_{t} \geq 0 \} > c \Bigl) \leq \min\{1, \frac{|\cX|}{2} e^{ -c / 2}\}$ due to $ \Pr\Bigl(\xi^2_{t} \mathbbm{1}\{ \xi_{t} \geq 0 \} > c \Bigl) \leq 1$, we obtain
    \begin{align*}
        \eqref{eq:acq_square_bound}
        &\leq \int^{2 \log (|\cX| / 2)}_{0} 1 {\rm d}c + \int^{\infty}_{2 \log (|\cX| / 2)} \frac{|\cX|}{2} e^{ -c / 2} {\rm d}c \\
        &= 2 \log (|\cX| / 2) + \frac{|\cX|}{2} \int^{\infty}_{0} e^{ - (c + 2 \log (|\cX| / 2)) / 2} {\rm d}c \\
        &= 2 \log (|\cX| / 2) + \int^{\infty}_{0} e^{ - c / 2} {\rm d}c \\
        &= 2 \log (|\cX| / 2) + 2,
    \end{align*}
    which concludes the proof.
\end{proof}

\subsection{Discrete Domain}
\label{app:PIMS_discrete}

For the discrete input domain, using Lemma~\ref{lem:bound_PIMS}, we show the following theorem:
\begin{reptheorem}{theo:BCR_PIMS_discrete}
    Let $f \sim \cG \cP (0, k)$, where $k$ is a stationary kernel and $k(\*x, \*x) = 1$, and $\cX$ be finite.
    Then, by running PIMS, BCR can be bounded as follows: 
    \begin{align*}
        {\rm BCR}_T \leq \sqrt{C_1 C_2 T \gamma_T},
    \end{align*}
    where $C_1 \coloneqq 2 / \log(1 + \sigma^{-2})$ and $C_2 \coloneqq 2 + 2 \log \bigl(|\cX|/2\bigr)$.
\end{reptheorem}
\begin{proof}
    Since $\*x_t \indep f | \cD_{t-1}$ and $\*x_t | \cD_{t-1}$ depends on only $g^*_t$, we can transform BCR as follows:
    \begin{align*}
        {\rm BCR}_T 
        &= \sum_{t=1}^T \EE_{\cD_{t-1}}\biggl[ \EE_t \bigl[ f(\*x^*) - f(\*x_t) \bigr] \biggr] \\
        &= \sum_{t=1}^T \EE_{\cD_{t-1}} \biggl[ \EE_t \bigl[ f(\*x^*) - g^*_t + g^*_t - f(\*x_t) \bigr] \biggr] \\
        &= \sum_{t=1}^T \EE_{\cD_{t-1}}\biggl[ \EE_t \bigl[ g^*_t - f(\*x_t) \bigr] \biggr] \mybecause{\EE_t[f(\*x^*)] = \EE_t [g^*_t]} \\
        &= \sum_{t=1}^T \EE_{\cD_{t-1}}\biggl[ \EE_t \bigl[ g^*_t - \mu_{t-1}(\*x_t)\bigr] \biggr] \mybecause{f \indep \*x_t | \cD_{t-1}} \\
        &= \sum_{t=1}^T \EE_{\cD_{t-1}}\biggl[ \EE_t \biggl[ \frac{g^*_t - \mu_{t-1}(\*x_t)}{\sigma_{t-1}(\*x_t)} \sigma_{t-1}(\*x_t) \biggr] \biggr].
    \end{align*}
    Then, by defining $\xi_{t} \coloneqq \frac{g^*_t - \mu_{t-1}(\*x_t)}{\sigma_{t-1}(\*x_t)}$, we can obtain
    \begin{align*}
        {\rm BCR}_T 
        &= \sum_{t=1}^T \EE_{\cD_{t-1}}\biggl[ \EE_t\biggl[ \xi_{t} \sigma_{t-1}(\*x_t) \biggr] \biggr] \\
        &= \EE\biggl[ \sum_{t=1}^T \xi_{t} \sigma_{t-1}(\*x_t) \biggr] \\
        &\leq \EE\biggl[ \sum_{t=1}^T \xi_{t} \mathbbm{1}\{ \xi_{t} \geq 0 \} \sigma_{t-1}(\*x_t) \biggr] \\
        &\leq \EE\biggl[ \sqrt{\sum_{t=1}^T \xi^2_{t} \mathbbm{1}\{ \xi_{t} \geq 0 \} \sum_{t=1}^T \sigma^2_{t-1}(\*x_t)} \biggr]. \mybecause{\text{Cauchy-Schwarz inequality}}
    \end{align*}
    We can use the following well-known bound of MIG:
    \begin{align*}
        \sum_{t=1}^T \sigma^2_{t-1}(\*x_t)
        &\leq C_1 \gamma_T, && \mybecause{\text{Lemma~5.4 in \citep{Srinivas2010-Gaussian}}}
    \end{align*}
    where $C_1 \coloneqq 2 / \log(1 + \sigma^{-2})$.
    Therefore, 
    \begin{align*}
        {\rm BCR}_T &\leq \EE\left[ \sqrt{\sum_{t=1}^T \xi^2_{t}\mathbbm{1}\{ \xi_{t} \geq 0 \}} \right] \sqrt{C_1 \gamma_T} \\
        &\leq \sqrt{ \EE\Bigl[\sum_{t=1}^T \xi^2_{t}\mathbbm{1}\{ \xi_{t} \geq 0 \} \Bigr]} \sqrt{C_1 \gamma_T} \mybecause{\text{Jensen inequality}} \\
        &= \sqrt{ \sum_{t=1}^T \EE\Bigl[\xi^2_{t}\mathbbm{1}\{ \xi_{t} \geq 0 \} \Bigr]} \sqrt{C_1 \gamma_T}.
    \end{align*}
    We can apply Lemma~\ref{lem:bound_PIMS} to $\xi_t$ for all $t \geq 1$.
    Thus, we obtain the following upper bound of BCR:
    \begin{align*}
        {\rm BCR}_T
        &\leq \sqrt{ \sum_{t=1}^T \EE\Bigl[\xi^2_{t}\mathbbm{1}\{ \xi_{t}(\*x_t) \geq 0 \} \Bigr]} \sqrt{C_1 \gamma_T} \\
        &\leq \sqrt{ C_1 C_2 T \gamma_T},
    \end{align*}
    where $C_2 \coloneqq 2 + 2 \log (|\cX| / 2)$.
\end{proof}

\subsection{Continuous Domain with Continuous Sample Path}
\label{app:PIMS_continuous}

\begin{reptheorem}{theo:BCR_PIMS_continuous}
    Let $f \sim \cG \cP (0, k)$, where $k$ is a stationary kernel, $k(\*x, \*x) = 1$, and Assumption~\ref{assump:continuous_X} holds.
    Then, by running PIMS, BCR can be bounded as follows:
    \begin{align*}
        {\rm BCR}_T \leq \frac{\pi^2}{6} + \sqrt{ C_1 T \gamma_T m_T},
    \end{align*}
    where $C_1 \coloneqq 2 / \log(1 + \sigma^{-2})$ and $m_t \coloneqq 2d \log \left( \left\lceil \! t^2 b d r \bigl( \log (ad) + \sqrt{\pi} / 2 \bigr) \sqrt{(\sigma^2 + n_t) / \sigma^2} \! \right\rceil \right) - 2 \log 2 + 2$.
\end{reptheorem}
\begin{proof}
    For the sake of analysis, we used a set of discretization $\cX_t \subset \cX$ for $t \geq 1$.
    For any $t \geq 1$, let $\cX_t \subset \cX$ be a finite set with each dimension equally divided into $\tau_t = \lceil t^2 b d r \bigl( \log (ad) + \sqrt{\pi} / 2 \bigr) \sqrt{(\sigma^2 + n_t) / \sigma^2} \rceil$.
    Thus, $|\cX_t| = \tau_t^d$.
    In addition, we define $[\*x]_t$ as the nearest point in $\cX_t$ of $\*x \in \cX$.
    Note that the discretization does not depend on any randomness and is deterministic.

    As with Theorem~\ref{theo:BCR_PIMS_discrete}, we obtain
    \begin{align*}
        {\rm BCR}_T
            &= \sum_{t=1}^T \EE\biggl[ \EE_t \biggl[ \frac{g^*_t - \mu_{t-1}(\*x_t)}{\sigma_{t-1}(\*x_t)} \sigma_{t-1}(\*x_t) \biggr] \biggr].
    \end{align*}
    Let $\tilde{g}^*_t \coloneqq \max_{\*x \in \cX_t} g_t (\*x)$, $\tilde{\*x}_t \coloneqq \argmin_{\*x \in \cX_t} \left\{ (\tilde{g}^*_t - \mu_{t-1}(\*x)) / \sigma_{t-1}(\*x) \right\}$, and $\*z^*_t \coloneqq \argmax_{\*x \in \cX} g_t (\*x)$.
    Then, we can obtain
    \begin{align*}
        {\rm BCR}_T
        &\leq \sum_{t=1}^T \EE\biggl[ \EE_t \biggl[ \frac{g^*_t  - \mu_{t-1}(\tilde{\*x}_t)}{\sigma_{t-1}(\tilde{\*x}_t)} \sigma_{t-1}(\*x_t) \biggr] \biggr] \mybecause{\text{$\*x_t$ is minimum}} \\
        &= \sum_{t=1}^T \EE\biggl[ \EE_t \biggl[ \frac{g^*_t - g([\*z^*_t]_t) + g([\*z^*_t]_t) - \tilde{g}^*_t + \tilde{g}^*_t- \mu_{t-1}(\tilde{\*x}_t)}{\sigma_{t-1}(\tilde{\*x}_t)} \sigma_{t-1}(\*x_t) \biggr] \biggr] \\
        &\leq \sum_{t=1}^T \EE\biggl[ \EE_t \biggl[ \frac{g^*_t - g([\*z^*_t]_t) + \tilde{g}^*_t- \mu_{t-1}(\tilde{\*x}_t)}{\sigma_{t-1}(\tilde{\*x}_t)} \sigma_{t-1}(\*x_t) \biggr] \biggr] \mybecause{ g([\*z^*_t]_t) \leq \tilde{g}^*_t} \\
        &= \sum_{t=1}^T \EE\biggl[ \EE_t \biggl[ \frac{g^*_t - g([\*z^*_t]_t) }{\sigma_{t-1}(\tilde{\*x}_t)} \sigma_{t-1}(\*x_t) + \frac{\tilde{g}^*_t - \mu_{t-1}(\tilde{\*x}_t)}{\sigma_{t-1}(\tilde{\*x}_t)} \sigma_{t-1}(\*x_t) \biggr] \biggr] \\
        &= \underbrace{\sum_{t=1}^T \EE\biggl[ \frac{g^*_t - g([\*z^*_t]_t) }{\sigma_{t-1}(\tilde{\*x}_t)} \sigma_{t-1}(\*x_t) \biggr]}_{\eqqcolon B_1}
        + \underbrace{\sum_{t=1}^T \EE\biggl[ \frac{\tilde{g}^*_t - \mu_{t-1}(\tilde{\*x}_t)}{\sigma_{t-1}(\tilde{\*x}_t)} \sigma_{t-1}(\*x_t) \biggr]}_{\eqqcolon B_2}
    \end{align*}
    For $B_2$, as with the proof of Theorem~\ref{theo:BCR_PIMS_discrete},
    \begin{align*}
        B_2 \leq \sqrt{C_1 \gamma_T} \sqrt{\sum_{t=1}^T \EE \left[ \tilde{\xi}_t^2 \mathbbm{1} \{\tilde{\xi}_t \geq 0\} \right]},
    \end{align*}
    where
    \begin{align*}
        \tilde{\xi}_t
        \coloneqq \frac{\tilde{g}^*_t - \mu_{t-1}(\tilde{\*x}_t)}{\sigma_{t-1}(\tilde{\*x}_t)}
        = \min_{\*x \in \cX_t}  \frac{\tilde{g}^*_t - \mu_{t-1}(\*x)}{\sigma_{t-1}(\*x)}.
    \end{align*}
    Hence, by replacing $\cX$ with $\cX_t$ compared with Theorem~\ref{theo:BCR_PIMS_discrete}, we can apply Lemma~\ref{lem:bound_PIMS} as follows:
    \begin{align*}
        \EE \left[ \tilde{\xi}_t^2(\tilde{\*x}_t) \mathbbm{1} \{\tilde{\xi}_t(\tilde{\*x}_t) \geq 0\} \right]
        &\leq 2 \log(|\cX_t| / 2) + 2 \\
        &= 2d \log \bigl( \left\lceil t^2 b d r \left( \log (ad) + \sqrt{\pi} / 2 \bigr) \sqrt{(\sigma^2 + n_t) / \sigma^2} \right\rceil \right) - 2 \log 2 + 2 \\
        &= m_t.
    \end{align*}
    Therefore, we obtain $B_2 \leq \sqrt{C_1 T \gamma_T m_T}$.

    Next, for $B_1$,
    \begin{align*}
        B_1
        &= \sum_{t=1}^T \EE\biggl[ \frac{g^*_t - g([\*z^*_t]_t) }{\sigma_{t-1}(\tilde{\*x}_t)} \sigma_{t-1}(\*x_t) \biggr] \\
        &\leq \sum_{t=1}^T \EE\biggl[ \frac{g^*_t - g([\*z^*_t]_t) }{\sigma_{t-1}(\tilde{\*x}_t)} \biggr] \mybecause{\sigma_{t-1}(\*x_t) \leq 1} \\
        &\leq \sum_{t=1}^T \EE\biggl[ \bigl( g^*_t - g([\*z^*_t]_t) \bigr) \sqrt{\frac{ \sigma^2 + n_t }{\sigma^2} } \biggr] \mybecause{\text{Lemma~\ref{lem:LB_posterior_variance}}} \\
        &\leq \sum_{t=1}^T \frac{1}{t^2} \mybecause{\text{Lemma~\ref{lem:discretized_error_sample_path}}} \\
        &\leq \frac{\pi^2}{6}.
    \end{align*}
    Consequently, we obtain the desired result.
\end{proof}

Finally, we show the BSR bound:
\begin{reptheorem}{theo:BSR_PIMS_continuous}
    Assume the same condition as in Theorem~\ref{theo:BCR_PIMS_continuous}.
    Then, by running PIMS, BSR can be bounded as follows:
    \begin{align*}
        {\rm BSR}_T \leq \frac{\pi^2}{6 T} + \sqrt{ \frac{C_1 \gamma_T m_T}{T}},
    \end{align*}
    where $C_1 \coloneqq 2 / \log(1 + \sigma^{-2})$ and $m_t \coloneqq 2d \log \left( \left\lceil \! t^2 b d r \bigl( \log (ad) + \sqrt{\pi} / 2 \bigr) \right\rceil \right) - 2 \log 2 + 2$.
\end{reptheorem}
\begin{proof}
    For the sake of analysis, we used a set of discretization $\cX_t \subset \cX$ for $t \geq 1$.
    For any $t \geq 1$, let $\cX_t \subset \cX$ be a finite set with each dimension equally divided into $\tau_t = \lceil t^2 b d r \bigl( \log (ad) + \sqrt{\pi} / 2 \bigr) \rceil$.
    Thus, $|\cX_t| = \tau_t^d$.
    In addition, we define $[\*x]_t$ as the nearest point in $\cX_t$ of $\*x \in \cX$.
    Note that the discretization does not depend on any randomness and is deterministic.

    Let $\tilde{g}^*_t \coloneqq \max_{\*x \in \cX_t} g_t (\*x)$, $\tilde{\*x}_t \coloneqq \argmin_{\*x \in \cX_t} \left\{ (\tilde{g}^*_t - \mu_{t-1}(\*x)) / \sigma_{t-1}(\*x) \right\}$, and $\*z^*_t \coloneqq \argmax_{\*x \in \cX} g_t (\*x)$.
    From Lemma~\ref{lem:BSR_bound}, BSR can be transformed as follows:
    \begin{align*}
        {\rm BSR}_T 
        &\leq \frac{1}{T} \sum_{t=1}^T {\rm BSR}_t \\
        &= \frac{1}{T} \sum_{t=1}^T \EE \left[ f(\*x^*) - f(\hat{\*x}_t) \right] \\
        &= \frac{1}{T} \sum_{t=1}^T \EE \left[ \EE_t \left[ f(\*x^*) \right] - \mu_{t-1}(\hat{\*x}_t) \right] \\
        &= \frac{1}{T} \sum_{t=1}^T \EE \left[ f(\*x^*) - f([\*x^*]_t) + f([\*x^*]_t) - g_t([\*z^*_t]_t) + g_t([\*z^*_t]_t) - \tilde{g}^*_t + \tilde{g}^*_t - \mu_{t-1}(\hat{\*x}_t) \right] \\
        &\leq \frac{\pi^2}{6T} + \frac{1}{T} \sum_{t=1}^T \EE \left[ \tilde{g}^*_t - \mu_{t-1}(\hat{\*x}_t) \right],
    \end{align*}
    where the final inequality can be obtained by the same proof of Theorem~\ref{theo:BCR_TS_continuous} and $ g_t([\*z^*_t]_t) \leq \tilde{g}^*_t $.

    Then, we show the inequality between $\sigma_{t-1}(\*x_t)$ and $\sigma_{t-1}(\tilde{\*x}_t)$.
    From the definition of $\*x_t$ and $\tilde{\*x}_t$, 
    \begin{align*}
        \frac{\tilde{g}^*_t - \mu_{t-1}(\*x_t)}{\sigma_{t-1}(\*x_t)} + \frac{g^*_t - \tilde{g}^*_t}{\sigma_{t-1}(\*x_t)} 
        = \frac{g^*_t - \mu_{t-1}(\*x_t)}{\sigma_{t-1}(\*x_t)}
        \leq \frac{g^*_t - \mu_{t-1}(\tilde{\*x}_t)}{\sigma_{t-1}(\tilde{\*x}_t)} = \frac{\tilde{g}^*_t - \mu_{t-1}(\tilde{\*x}_t)}{\sigma_{t-1}(\tilde{\*x}_t)} + \frac{g^*_t - \tilde{g}^*_t}{\sigma_{t-1}(\tilde{\*x}_t)},
    \end{align*}
    and
    \begin{align*}
        \frac{\tilde{g}^*_t - \mu_{t-1}(\tilde{\*x}_t)}{\sigma_{t-1}(\tilde{\*x}_t)}
        \leq \frac{\tilde{g}^*_t - \mu_{t-1}(\*x_t)}{\sigma_{t-1}(\*x_t)}.
    \end{align*}
    Therefore, by combining the above two inequalities and $g^*_t \geq \tilde{g}^*_t$, we see that
    \begin{align*}
        \sigma_{t-1}(\tilde{\*x}_t) \leq \sigma_{t-1}(\*x_t).
    \end{align*}

    Hence, we can bound the remained term as follows:
    \begin{align*}
        \frac{1}{T} \sum_{t=1}^T \EE \left[ \tilde{g}^*_t - \mu_{t-1}(\hat{\*x}_t) \right] 
        &= \frac{1}{T} \sum_{t=1}^T \EE \left[ \frac{\tilde{g}^*_t - \mu_{t-1}(\hat{\*x}_t)}{\sigma_{t-1}(\*x_t)} \sigma_{t-1}(\*x_t) \right] \\
        &\leq \frac{1}{T} \sum_{t=1}^T \EE \left[ \frac{\tilde{g}^*_t - \mu_{t-1}(\hat{\*x}_t)}{\sigma_{t-1}(\*x_t)} \mathbbm{1}\{ \tilde{g}^*_t - \mu_{t-1}(\hat{\*x}_t) \} \sigma_{t-1}(\*x_t) \right] \\
        &\leq \frac{1}{T} \sum_{t=1}^T \EE \left[ \frac{\tilde{g}^*_t - \mu_{t-1}(\hat{\*x}_t)}{\sigma_{t-1}(\tilde{\*x}_t)} \mathbbm{1}\{ \tilde{g}^*_t - \mu_{t-1}(\hat{\*x}_t) \} \sigma_{t-1}(\*x_t) \right] && \bigl(\because \sigma_{t-1}(\tilde{\*x}_t) \leq \sigma_{t-1}(\*x_t) \bigr) \\
        &\leq \frac{1}{T} \sum_{t=1}^T \EE \left[ \frac{\tilde{g}^*_t - \mu_{t-1}(\tilde{\*x}_t)}{\sigma_{t-1}(\tilde{\*x}_t)} \mathbbm{1}\{ \tilde{g}^*_t - \mu_{t-1}(\tilde{\*x}_t) \} \sigma_{t-1}(\*x_t) \right] && \bigl(\because \mu_{t-1}(\tilde{\*x}_t) \leq \mu_{t-1}(\hat{\*x}_t) \bigr) \\
        &\leq \sqrt{ \frac{C_1 \gamma_T m_T}{T}},
    \end{align*}
    where the final inequality can be obtained as with $B_2$ in the proof of Theorem~\ref{theo:BCR_PIMS_continuous}.
\end{proof}

\subsection{Continuous Domain with Discretized Sample Path}
\label{app:disretizedPIMS_continuous}

\begin{theorem}
    Let $f \sim \cG \cP (0, k)$, where $k$ is a stationary kernel, $k(\*x, \*x) = 1$, and Assumption~\ref{assump:continuous_X} holds.
    Then, by running PIMS that uses $\tilde{g}^*_t$ instead of $g^*_t$, i.e., evaluates $\tilde{\*x}_t = \argmin_{\*x \in \cX_t} (\tilde{g}^*_t - \mu_{t-1}(\*x)) / \sigma_{t-1}(\*x)$, BCR can be bounded as follows:
    \begin{align*}
        {\rm BCR}_T \leq \frac{\pi^2}{6} + \sqrt{ C_1 T \gamma_T m_T},
    \end{align*}
    where $C_1 \coloneqq 2 / \log(1 + \sigma^{-2})$ and $m_t \coloneqq 2d \log \left( \left\lceil \! t^2 b d r \bigl( \log (ad) + \sqrt{\pi} / 2 \bigr) \right\rceil \right) - 2 \log 2 + 2$.
\end{theorem}
\begin{proof}
    We can transform the BCR as follows:
    \begin{align*}
        {\rm BCR}_T
        &= \sum_{t=1}^T \EE\biggl[ \EE_t \biggl[ g^*_t - g([\*z^*_t]_t) + g([\*z^*_t]_t) - \tilde{g}^*_t + \tilde{g}^*_t  - \mu_{t-1}(\tilde{\*x}_t) \biggr] \biggr] \\
        &\leq \sum_{t=1}^T \EE\biggl[ \EE_t \biggl[ g^*_t - g([\*z^*_t]_t) + \tilde{g}^*_t  - \mu_{t-1}(\tilde{\*x}_t) \biggr] \biggr] \mybecause{\tilde{g}^*_t \geq g([\*z^*_t]_t)} \\
        &= \sum_{t=1}^T \EE\left[ g^*_t - g([\*z^*_t]_t) \right] 
        + \sum_{t=1}^T \EE \biggl[  \frac{\tilde{g}^*_t  - \mu_{t-1}(\tilde{\*x}_t)}{\sigma_{t-1}(\tilde{\*x}_t)} \sigma_{t-1}(\tilde{\*x}_t) \biggr],
    \end{align*}
    where $\*z^*_t = \argmax_{\*x \in \cX} g_t(\*x)$.
    Then, the first term can be bounded above by $\pi^2 / 6$ using Lemma~\ref{lem:discretized_error_sample_path}.
    Furthermore, the second term can be bounded above by $\sqrt{C_1 T \gamma_T m_T}$ as with the proof of Theorem~\ref{theo:BCR_PIMS_discrete}.
\end{proof}

\section{Auxiliary Lemmas}
\label{app:auxiliary_lemmas}

We used the following useful lemmas:
\begin{lemma}[in Lemma 5.2 of \citep{Srinivas2010-Gaussian} and Lemma~H.3 of \citep{Takeno2023-randomized}]
    For $c > 0$, the survival function of the standard normal distribution can be bounded above as follows:
    \begin{align*}
        1 - \Phi(c) \leq \frac{1}{2} \exp (- c^2 / 2).
    \end{align*}
    \label{lem:Gauss_tail_bound}
\end{lemma}
\begin{lemma}[Lemma~H.1 of \citep{Takeno2023-randomized}]
    Let $f \sim \cG \cP (0, k)$ and Assumption~\ref{assump:continuous_X} holds.
    Let the supremum of the partial derivatives $L_{\rm max} \coloneqq \sup_{j \in [d]} \sup_{\*x \in \cX} \left| \frac{\partial f}{\partial x_j} \right|$.
    Then, $\EE[L_{\rm max}]$ can be bounded above as follows:
    \begin{align*}
        \EE[L_{\rm max}] \leq b \bigl( \sqrt{\log (ad)} + \sqrt{\pi} / 2 \bigr).
    \end{align*}
    \label{lem:L_max_bound}
\end{lemma}

\begin{lemma}[Lemma~H.2 of \citep{Takeno2023-randomized}]
    Let $f \sim \cG \cP (0, k)$ and Assumption~\ref{assump:continuous_X} holds.
    Let $\cX_t \subset \cX$ be a finite set with each dimension equally divided into $\tau_t \geq bdr u_t \bigl( \sqrt{\log (ad)} + \sqrt{\pi} / 2 \bigr)$ for any $t \geq 1$.
    Then, we can bound the expectation of differences,
    \begin{align*}
        \sum_{t=1}^T \EE \left[ \sup_{\*x \in \cX} | f(\*x) - f( [\*x]_t ) |\right] \leq \sum_{t=1}^T \frac{1}{u_t},
    \end{align*}
    where $[\*x]_t$ is the nearest point in $\cX_t$ of $\*x \in \cX$.
    \label{lem:discretized_error_sample_path}
\end{lemma}

Similarly, we can obtain the following lemma:
\begin{lemma}
    Let $f \sim \cG \cP (0, k)$ and assume the same condition as in Lemma~\ref{lem:Lipschitz_std}.
    Let $\cX_t \subset \cX$ be a finite set with each dimension equally divided into $\tau_t \geq drL_{\sigma} u_t $ for any $t \geq 1$.
    Then, the following inequality holds with probability 1:
    \begin{align*}
        \forall t \geq 1, \forall \cD_{t-1}, \sup_{\*x \in \cX} | \sigma_{t-1}(\*x) - \sigma_{t-1}( [\*x]_t ) | \leq \frac{1}{u_t},
    \end{align*}
    where $[\*x]_t$ is the nearest point in $\cX_t$ of $\*x \in \cX$.
    \label{lem:discretized_error_std}
\end{lemma}
\begin{proof}
    From the construction of $\cX_t$, we can obtain the upper bound of L1 distance between $\*x$ and $[\*x]_t$ as follows:
    \begin{align}
        \sup_{\*x \in \cX}\| \*x - [\*x]_t \|_1
        &\leq \frac{dr }{dr L_{\sigma} u_t} \nonumber \\
        &= \frac{1}{L_{\sigma} u_t}.
        \label{eq:sup_L1dist_discretization_general}
    \end{align}
    Note that since this discretization is fixed beforehand, the discretization does not depend on any randomness.

    Then, we obtain the following:
    \begin{align*}
         \EE \left[ \sup_{\*x \in \cX} | \sigma_{t-1}(\*x) - \sigma_{t-1}( [\*x]_t ) | \right] 
        &\leq L_{\sigma} \sup_{\*x \in \cX} \| \*x - [\*x]_t \|_1 \mybecause{\text{Lemma~\ref{lem:Lipschitz_std}}} \\
        &\leq L_{\sigma} \frac{1}{L_{\sigma} u_t} && \bigl(\because \text{Eq.~\eqref{eq:sup_L1dist_discretization_general}} \bigr) \\
        &\leq \frac{1}{u_t}. && \bigl(\because \text{Lemma~\ref{lem:L_max_bound}} \bigr)
    \end{align*}
\end{proof}

For BSR, based on the discussion in Appendix~A of \citet{Takeno2023-randomized}, we show the following lemma:
\begin{replemma}{lem:BSR_bound}
    BSR can be bounded from above as follows:
    \begin{align*}
        \overline{\rm BSR}_T 
        \leq \sum_{t=1}^T \overline{\rm BSR}_t / T
        \leq {\rm BCR}_T / T.
    \end{align*}
    and
    \begin{align*}
        {\rm BSR}_T
        \leq {\rm BCR}_T / T
    \end{align*}
\end{replemma}
\begin{proof}
    For the first inequality, we can bound the modified BSR from above as follows:
    \begin{align*}
        \overline{\rm BSR}_T 
        &= \EE \left[ f(\*x^*) - f(\hat{\*x}_T) \right] \\
        &= \EE_{\cD_{T-1}} \left[ \EE_T \bigl[ f(\*x^*) - f(\hat{\*x}_T) \bigr] \right] \\
        &= \EE_{\cD_{T-1}} \left[ \EE_T \bigl[ f(\*x^*) \bigr] - \mu_{T-1}(\hat{\*x}_T) \right] \\
        &\leq \EE_{\cD_{T-1}} \left[ \EE_T \left[ f(\*x^*) \right] - \frac{1}{T} \sum_{t=1}^T \mu_{T-1}(\hat{\*x}_t) \right] && \bigl(\because \forall t \leq T, \mu_{T-1}(\hat{\*x}_t) \leq \mu_{T-1}(\hat{\*x}_T) \bigr) \\
        &= \EE_{\cD_{T-1}} \left[ \EE_T \left[ f(\*x^*)  - \frac{1}{T} \sum_{t=1}^T f(\hat{\*x}_t) \right] \right] \\
        &= \frac{1}{T} \sum_{t=1}^T \EE \left[ f(\*x^*)  - f(\hat{\*x}_t) \right] \\
        &= \frac{1}{T} \sum_{t=1}^T \overline{\rm BSR}_t.
    \end{align*}
    In addition, we can obtain
    \begin{align*}
        \frac{1}{T} \sum_{t=1}^T \overline{\rm BSR}_t
        &= \frac{1}{T} \sum_{t=1}^T \EE \left[ f(\*x^*) - f(\*x_t) + f(\*x_t) - f(\hat{\*x}_t) \right] \\
        &= \EE_{\cD_{t-1}} \left[ \frac{1}{T} \sum_{t=1}^T \biggl\{ \EE_t \bigl[ f(\*x^*) - f(\*x_t) \bigr] + \mu_{t-1}(\*x_t) - \mu_{t-1}(\hat{\*x}_t) \biggr\} \right] \\
        &\leq \frac{1}{T} \sum_{t=1}^T \EE \left[ f(\*x^*)  - f(\*x_t) \right] &&\bigl(\because \forall t \geq 1, \mu_{t-1}(\*x_t) \leq \mu_{t-1}(\hat{\*x}_t) \bigr) \\
        &= \frac{1}{T} {\rm BCR}_T.
    \end{align*}
    The inequality for BSR can be obtained from the relationship between average and the maximum $\max_{t \leq T} f(\*x_t) \geq \sum_{t=1}^T f(\*x_t) / T$.
\end{proof}

\begin{lemma}
    Suppose that $\cX$ is a discrete input domain.
    Let $\xi_t = \min_{\*x \in \cX} \frac{g^*_t - \mu_{t-1}(\*x)}{\sigma_{t-1}(\*x)} = \frac{g^*_t - \mu_{t-1}(\*x_t)}{\sigma_{t-1}(\*x_t)}$.
    Then, for all $t \in \NN$ and $c \geq 0$,
    \begin{align}
        \Pr\bigl( \xi_t > c \bigr) = \Pr \bigl( g^*_t > \max_{\*x \in \cX} \left\{ \mu_{t-1}(\*x) + c \sigma_{t-1} (\*x) \right\} \bigr).
    \end{align}
    \label{lem:acq_tail_bound}
\end{lemma}
\begin{proof}
    From Lemma~\ref{lem:equivalence_PIMS}, $\*x_t \in \argmax_{\*x \in \cX} \left\{ \mu_{t-1}(\*x) + \frac{g^*_t - \mu_{t-1}(\*x_t)}{\sigma_{t-1}(\*x_t)} \sigma_{t-1}(\*x) \right\}$.
    Therefore, if $\xi_t > c $, i.e.,
    \begin{align*}
        \frac{g^*_t - \mu_{t-1}(\*x_t)}{\sigma_{t-1}(\*x_t)} > c,
    \end{align*}
    then
    \begin{align*}
        g^*_t 
        &= \max_{\*x \in \cX} \left\{ \mu_{t-1}(\*x) + \frac{g^*_t - \mu_{t-1}(\*x_t)}{\sigma_{t-1}(\*x_t)} \sigma_{t-1} (\*x)\right\} \\
        &> \max_{\*x \in \cX} \left\{ \mu_{t-1}(\*x) + c \sigma_{t-1} (\*x)\right\}.
    \end{align*}
    Inversely, if 
    \begin{align*}
        g^*_t 
        &= \max_{\*x \in \cX} \left\{ \mu_{t-1}(\*x) + \frac{g^*_t - \mu_{t-1}(\*x_t)}{\sigma_{t-1}(\*x_t)} \sigma_{t-1} (\*x)\right\} \\
        &> \max_{\*x \in \cX} \left\{ \mu_{t-1}(\*x) + c \sigma_{t-1} (\*x)\right\}.
    \end{align*}
    then 
    \begin{align*}
        \frac{g^*_t - \mu_{t-1}(\*x_t)}{\sigma_{t-1}(\*x_t)} > c.
    \end{align*}
    Hence, the statement of the lemma holds.
\end{proof}

Finally, we show the following lemma for completeness:
\begin{replemma}{lem:LB_posterior_variance}
    Let $k$ be a kernel s.t. $k(\*x, \*x) = 1$.
    %
    Then, the posterior variance is bounded from below as,
    \begin{align}
        \sigma^2_{t}(\*x) \geq \frac{\sigma^2}{\sigma^2 + n_t},
    \end{align}
    for all $\*x \in \cX$ and for all $t \geq 0$.
\end{replemma}
\begin{proof}
    When only one input $\*x$ is repeatedly evaluated $t$ times, $\sigma_{t}(\*x)$ become the minimum.
    For simplicity, let $\cD_0 = \emptyset$ without loss of generality.
    Then, we see that
    \begin{align*}
        \sigma^2_{0}(\*x) &= 1, \\
        \sigma^2_{1}(\*x) 
        &= \frac{ \sigma^2}{1 + \sigma^2}.
    \end{align*}
    Assume $\sigma^2_{i}(\*x) = \frac{ \sigma^2}{i + \sigma^2}$.
    Then, we can obtain
    \begin{align*}
        \sigma^2_{i+1}(\*x) 
        &= \sigma^2_{i}(\*x) - \frac{ \left( \sigma^2_{i}(\*x) \right)^2}{\sigma^2_{i}(\*x) + \sigma^2} \\
        &= \frac{ \sigma^2 \sigma^2_{i}(\*x)}{\sigma^2_{i}(\*x) + \sigma^2} \\
        &= \frac{ \frac{\sigma^4}{i + \sigma^2}}{\frac{(i+1)\sigma^2 + \sigma^4}{i + \sigma^2}} \\
        &= \frac{ \sigma^2}{(i+1) + \sigma^2}.
    \end{align*}
    Since $\sigma^2_{i}(\*x) = \frac{ \sigma^2}{i + \sigma^2}$ holds when $i=0$ and $1$, the statement $\sigma^2_{t}(\*x) = \frac{ \sigma^2}{t + \sigma^2}$ holds for all $t \geq 0$.
    If $|\cD_0| = n_0 > 0$, $t$ changes to $n_t = n_0 + t$.
\end{proof}

\end{document}